\documentclass[11pt]{article}

\usepackage{acl}

\usepackage{times}
\usepackage{latexsym}

\usepackage[T1]{fontenc}
\usepackage[utf8]{inputenc}

\usepackage{microtype}

\usepackage{inconsolata}

\usepackage{graphicx}
\usepackage{amsmath}
\usepackage{multirow}
\usepackage{booktabs}
\usepackage[table]{xcolor}
\definecolor{rowgray}{gray}{0.90}
\usepackage{amssymb}
\title{BiMind: A Dual-Head Reasoning Model with Attention-Geometry Adapter for Incorrect Information Detection}

\author{
  Zhongxing Zhang\textsuperscript{1}\textsuperscript{,}\thanks{Corresponding authors}, Emily K. Vraga\textsuperscript{1}, Jisu Huh\textsuperscript{1}, Jaideep Srivastava\textsuperscript{1}\textsuperscript{,}\footnotemark[1]\\
  \textsuperscript{1}University of Minnesota, Twin Cities \\
  \texttt{\{zhan8889, ekvraga, jhuh, srivasta\}@umn.edu}
}

\begin{document}
\maketitle
\begin{abstract}
Incorrect information poses significant challenges by disrupting content veracity and integrity, yet most detection approaches struggle to jointly balance textual content verification with external knowledge modification under collapsed attention geometries. To address this issue, we propose a dual-head reasoning framework, \textbf{BiMind}, which disentangles \emph{content-internal reasoning} from \emph{knowledge-augmented reasoning}. In BiMind, we introduce three core innovations: (i) an \textbf{attention geometry adapter} that reshapes attention logits via token-conditioned offsets and mitigates attention collapse; (ii) a \textbf{self-retrieval knowledge mechanism}, which constructs an in-domain semantic memory through kNN retrieval and injects retrieved neighbors via feature-wise linear modulation; (iii) the \textbf{uncertainty-aware fusion strategies}, including entropy-gated fusion and a trainable agreement head, stabilized by a symmetric Kullback-Leibler agreement regularizer. To quantify the knowledge contributions, we define a novel metric, \textbf{Value-of-eXperience (VoX)}, to measure instance-wise logit gains from knowledge-augmented reasoning. Experiment results on public datasets demonstrate that our BiMind model outperforms advanced detection approaches and provides interpretable diagnostics on when and why knowledge matters.
Our BiMind model and the tested datasets are available at \href{https://github.com/cvzh/BiMind}{https://github.com/cvzh/BiMind}.
% Our findings highlight the benefits of modeling the interplay between content reasoning and knowledge reasoning in incorrect information detection.
\end{abstract}

\begin{figure}
  \includegraphics[width=\columnwidth]{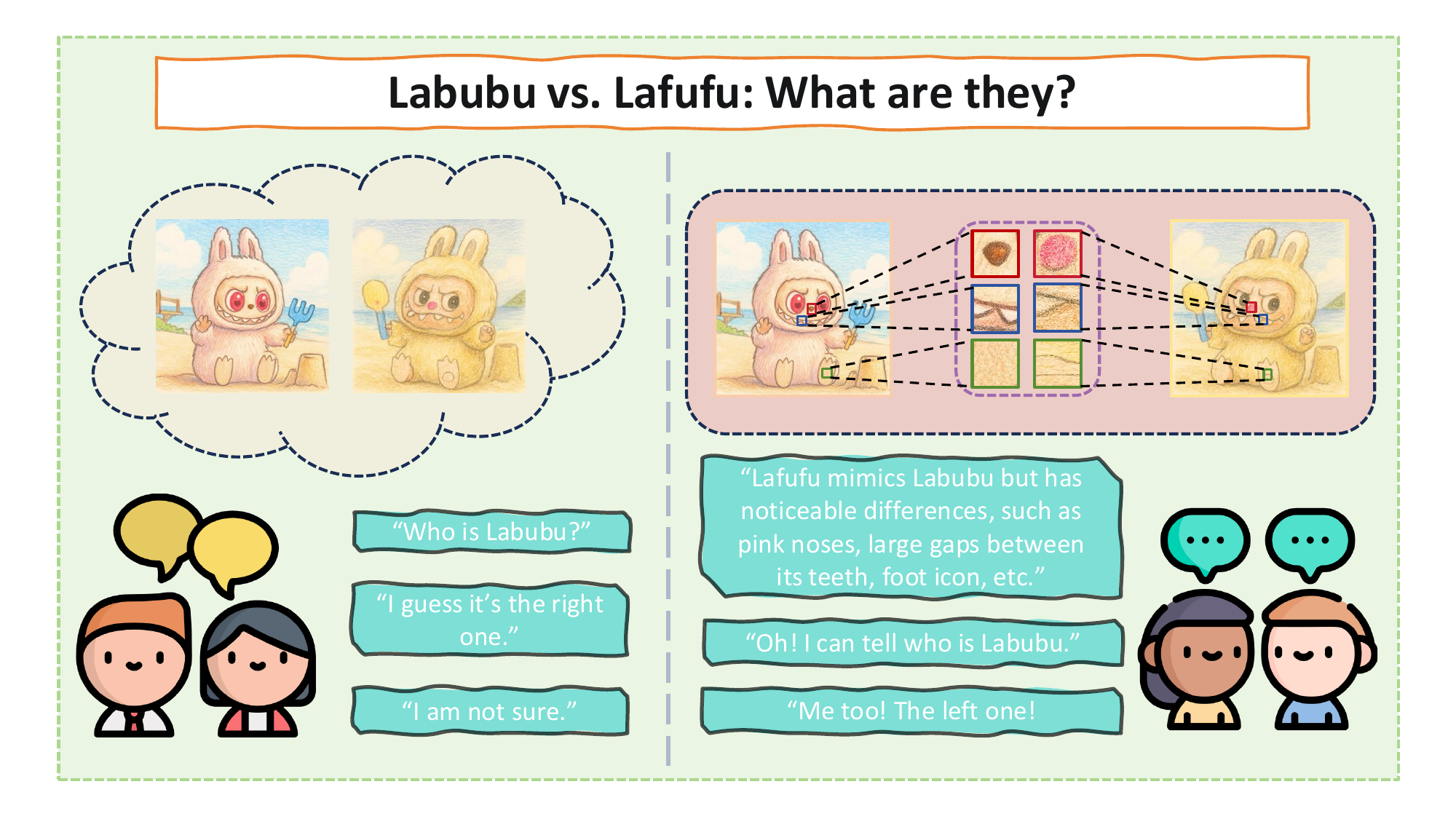}
  \caption{Illustrative case of self-correction via knowledge: without knowledge (left), the content-internal head labels Lafufu (counterfeit Labubu) as Labubu; with knowledge (right), the knowledge-augmented head corrects the label as Lafufu.}
  \label{fig:self_correction}
\end{figure}

% \begingroup
% \renewcommand\thefootnote{}
% \footnotetext{*}
% \endgroup

\section{Introduction}
Nowadays, with the rapid rise of social media platforms, such as X (Twitter), Instagram, and TikTok, an increasing number of individuals or communities rely on these online platforms for communication, information dissemination, and education, especially during the pandemic \citep{tsao2021social}. Though the conveniences brought by social media, the content correctness (i.e., factual accuracy and alignment with verifiable evidence) of information disseminated still falls short of media standards and social expectations, compared to traditional media platforms, e.g., television and newspapers \citep{shu2017fake, zhou2020survey}. A large volume of unverified or distorted content is easily produced and propagated through social media platforms \citep{ahmed2022combining}. Given that such incorrect information (e.g., spam \citep{wang-etal-2016-learning-represent}, rumor \citep{Bian2020RumorDO}, etc.) has significant negative impacts on individuals and society, such as social trust and information credibility \citep{thorson2010credibility, bhattarai2021explainable, mazzeo2021detection, lan2025contextual}, addressing incorrect information propagation has become crucial in the areas of social media, mass communication, and public health. Technically, automatic models are developed to identify and detect the incorrect information on social media platforms, thereby mitigating the social effects \citep{10.1145/3393880, yang-etal-2023-rumor, shi-etal-2023-multiview}.

While incorrect information detection methods have achieved significant advances, these methods still struggle with feature complexity, knowledge injection, and attention collapse. Specifically, prior work focuses either on textual content features (e.g., linguistic features and contextual embeddings) or on external knowledge (knowledge graphs and retrieval-augmented generation), which integrates all the feature streams into a classifier, without any disentanglement between what is learned from textual content and what is contributed by external knowledge. As illustrated in Figure~\ref{fig:self_correction}, without knowledge inputs, the reader potentially accepts the incorrect information (i.e., Lafufu) as correct (i.e., Labubu) from the raw content; once the reader obtains relevant knowledge, the information is identified as incorrect. 

To uncover the interplay between content reasoning and knowledge reasoning, we propose a new view: disentangling content reasoning from knowledge reasoning in an explicit and structured way. In this paper, we introduce a novel dual-head model architecture, \textbf{BiMind}, for incorrect information detection. Our model employs two separate heads to explore content and knowledge features, respectively, where the knowledge is retrieved from an in-domain memory. This separation mechanism allows us to measure, analyze, and apply the two streams of features in a structured way. Technically, our three contributions drive BiMind's novelty:
\begin{itemize}
  % \item First, we introduce a behavior-aware encoding adapter, inspired by the observation that verbs are normally indicators of stance, uncertainty, or manipulation in incorrect information.
  \item First, we introduce an attention geometry adapter (AGA) that reshapes attention distributions at the pre-softmax logit level, stabilizing text encoding by mitigating attention collapse.
  \item Second, we design a self-retrieval knowledge module that encodes the training set into an in-domain semantic memory and then injects nearest-neighbor features via feature-wise linear modulation (FiLM).
  \item Third, we propose two uncertainty-aware fusion strategies, i.e., entropy-gated fusion and a trainable agreement head, where we adapt a symmetric Kullback–Leibler (KL) regularizer to ensure consistency between heads.
  \item Finally, we define a novel metric, Value-of-eXperience (VoX), to quantify the contributions from external knowledge, improving model interpretability.
\end{itemize}
Experimental results demonstrate that our model enhances detection accuracy and interpretability, especially when external knowledge contributes to model predictions.

\section{Related Work}

\subsection{Content-based Methods}
Today, machine learning (ML) and natural language processing methods \citep{Kadhim2019SurveyOS, su2020motivations} have emerged as advanced tools to classify textual information in news articles into one or more predefined classes, such as correct or incorrect. Traditional ML methods, such as support vector machine, random forest, and decision tree, are commonly used in news content classification; however, these methods require hand-crafted features and struggle with complex text features, thus compromising performance \citep{Minaee2020DeepLT}.

Along with neural networks being boosted, deep learning frameworks have further enhanced the classification performance by extracting complex content features and capturing nuanced semantic features, such as convolutional neural networks (CNNs) \citep{Kim2014ConvolutionalNN, Wang2017LiarLP, kaliyar2020fndnet}, recurrent neural networks (RNNs) \citep{Ma2016DetectingRF, ruchansky2017csi}, and long short-term memory (LSTM) \citep{Sachan2019RevisitingLN, ma-etal-2020-mode}. Kaliyar et al. \citep{kaliyar2020fndnet} proposed a deep CNN model for incorrect information detection compared to classical CNN and LSTM structures, where it explores pre-trained word embeddings and multiple hidden layers to extract text features.

Additionally, attention networks integrated different features extracted from different latent aspects of news articles to improve detection accuracy \citep{Yang2016HierarchicalAN, linmei-etal-2019-heterogeneous, sun-lu-2020-understanding, yun2023focus}. For example, a hierarchical attention network (HAN) \citep{Yang2016HierarchicalAN} was proposed to capture hierarchical structure of documents and employ the word-level and sentence-level attentions. To construct structured graphs based on texts, graph convolutional networks (GCNs) \citep{Yao2018GraphCN, 10.1145/3714456} have been applied to textual content classification tasks, which construct document-level and corpus-level graphs to learn relationships among words, documents, and corpus.

With the aid of pre-trained knowledge embeddings, the transformer-based models have advanced the detection accuracy of incorrect information in news articles \citep{croce-etal-2020-gan, kaliyar2021fakebert, xiong-etal-2021-fusing, van-nooten-daelemans-2025-jump}. Combining the bidirectional encoder representations from transformers (BERT) \citep{devlin2019bert} with a CNN structure, Kaliyar et al. \citep{kaliyar2021fakebert} proposed a BERT-based incorrect information detection model, where it inputs the BERT embeddings into one-dimensional CNN layers and then detects incorrect information using local features and global dependencies. Along with the data structure and modality extending, multimodal approaches are proposed to handle more intricate detection tasks for incorrect information content across text, image, video, audio data, or multiple languages \citep{conneau2019cross, abdali2024multi, 10.1145/3637528.3671977, lu-koehn-2025-learn}. For instance, Wu et al. \citep{10.1145/3637528.3671977} emphasized the substantive content over stylistic features, using Large Language Models (LLMs) to reframe news articles and focus on content veracity. Though LLMs emerged with impressive capability of processing multimodal features, LLMs still require a large volume of data to update the known knowledge and maintain performance.

\subsection{Knowledge-based Methods}
Traditional detection methods focus on internal content features and external fact-checking resources to detect incorrect information \citep{vlachos2014fact, hassan2015detecting, guo2022survey}. For instance, the fact-checking approaches can identify and classify the texts by using the external knowledge sources to fact-check the news content \citep{etzioni2008open, wu2014toward, shi2016fact, vo2018rise}. But, these fact-checking approaches are time-consuming and demand human annotations, limiting the scalability and efficiency.

For further exploiting the content and external knowledge features to detect incorrect information, the credibility-based knowledge methods \citep{10.1145/3041021.3053379, 10.1145/3184558.3188731, Deng_Wang_Zhu_Wang_Feng_2025} were proposed, which extract the source and content credibility features to identify factual news from non-credible ones, thereby enhancing model performance.

To explore the user behavior, engagements, and interactions on social media, the social relationship-aware approaches \citep{10.1145/3274327, Shu2017BeyondNC, 10.1145/3404835.3462990, teng2022characterizing} were proposed, which can capture user relationships, news content, and dissemination patterns to improve detection accuracy. For instance, Shu et al. \citep{Shu2017BeyondNC} presented a tri-relationship-based detection framework of incorrect information content, where it explores the tri-relationship among publishers, news pieces, and users to differentiate reliable and unreliable articles. Zhang et al. \citep{Zhang2024HeterogeneousST} explored the heterogeneous subgraph transformer (HeteroSGT) to detect incorrect information via the heterogeneous graph by unearthing the relationships among news topics, entities, and content.

To understand the propagation patterns of incorrect information within social networks, the network-based methods \citep{Zhou2019NetworkbasedFN} were suggested, where these methods focus on the interactions among spreaders and their influence on information propagation. Ma et al. \citep{Ma2018RumorDO} presented tree-structured recursive neural networks to model the propagation pattern of tweets for detecting rumors on social media. Typically, graph-based approaches were proposed \citep{Bian2020RumorDO, Fu2022DISCOCA} to explore the potential of graph structure in modeling social context structures, including knowledge-driven \citep{Wang2018EANNEA, dun2021kan}, propagation-based \citep{Zhu2024PropagationSG}, and context-aware approaches \citep{10.1145/3589334.3648152, li-etal-2025-semantic}.

Another direction of incorrect information detection approaches focuses on enhancing model performance with knowledge generation. Retrieval-augmented methods \citep{guu2020retrieval, lewis2020rag, chen2026actormindemulatinghumanactor} apply nearest-neighbor retrieval into LLMs to improve factual reasoning. Though achieving expected performance, these methods are computationally intensive and entangle retrieved knowledge with raw content in an opaque way. In contrast, our model, BiMind, disentangles {content-internal reasoning} from {knowledge-augmented reasoning} using a single yet transparent architecture. This separation strategy allows us to explicitly quantify the value of external knowledge through our proposed uncertainty-aware fusion and VoX metric, which differentiates our model from generic knowledge embedding frameworks.

\begin{figure*}[!t]
  \centering
  \includegraphics[width=2.12\columnwidth]{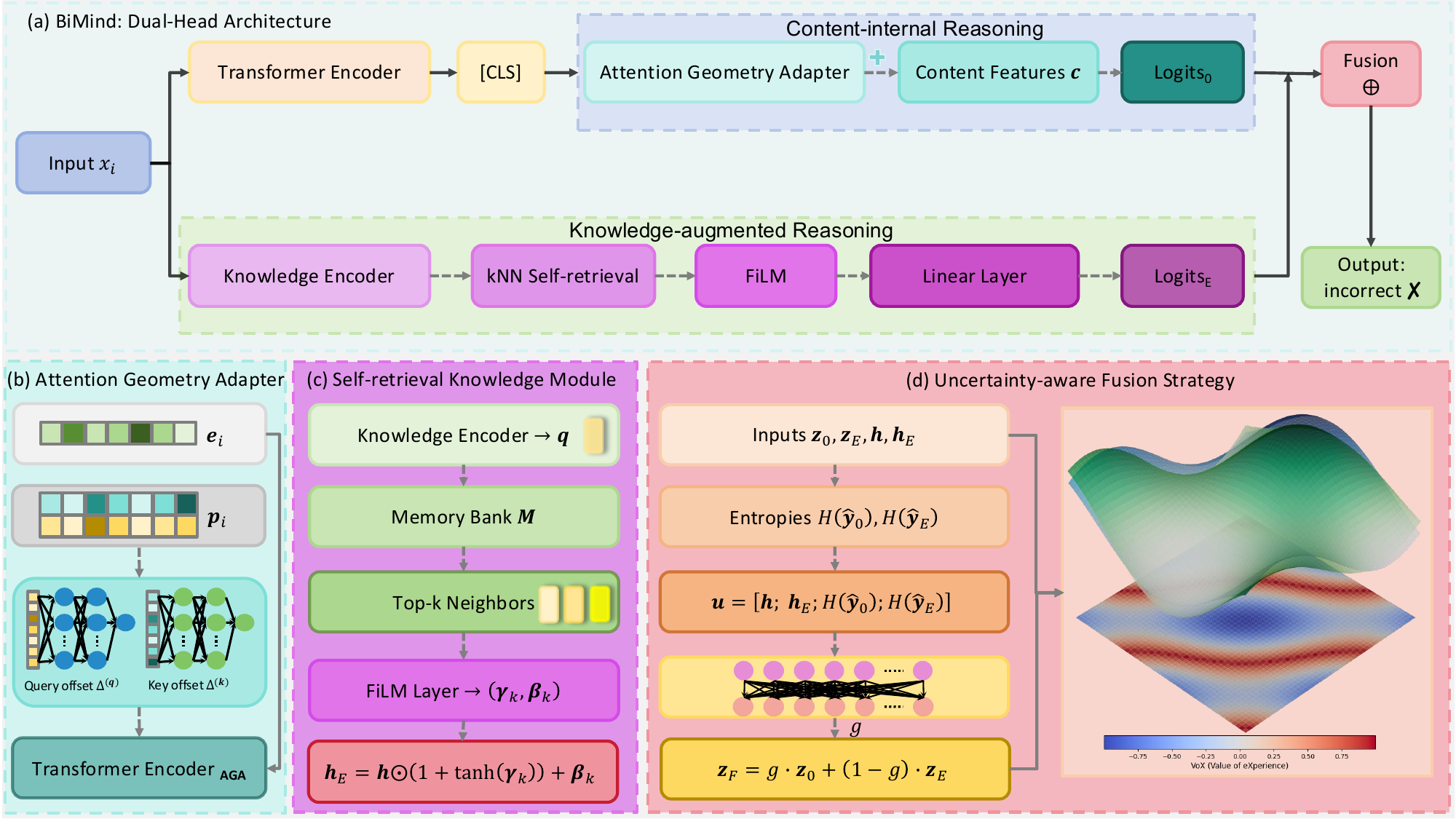}
  \caption{\textbf{An illustration of our proposed BiMind framework.} (a) Dual-head architecture with a content-internal head (top) and a knowledge-augmented head (bottom). (b) Attention geometry adapter reshapes pre-softmax attention logits via token-conditioned offsets. (c) Self-retrieval knowledge module retrieves top-$k$ neighbors and injects knowledge via FiLM to provide knowledge-augmented representations. (d) Uncertainty-aware fusion combines head logits via an entropy gating, with the VoX metric quantifying the knowledge contributions by comparing head outcomes, where blue and green surfaces in the instance space represent the content and knowledge reasoning heads, respectively.}
  \label{fig: pipeline}
\end{figure*}

\section{Methodology}
In this section, we introduce the fundamental framework of our proposed BiMind model, as shown in Figure~\ref{fig: pipeline}. Here, we define the raw input text \(x_i\) as an internal information unit; all auxiliary information beyond the raw content, such as that retrieved from in-domain memory or linked to external resources, is treated as external knowledge unit. Our objective is to disentangle \emph{content-internal reasoning} from \emph{knowledge-augmented reasoning} in a structured way, and to provide interpretable diagnostics on when and why external knowledge contributes to incorrect information detection. We present \textbf{BiMind}, a dual-head model with five key ingredients: (1) an attention geometry adapter (AGA) that reshapes pre-softmax attention geometry; (2) a self-retrieval knowledge module that constructs an in-domain memory through kNN retrieval; (3) a FiLM-based layer that injects retrieved external knowledge into the text representations; (4) the uncertainty-aware fusion strategies, including entropy-gated fusion strategy and a trainable agreement head, restrained by a symmetric KL regularizer; and (5) a VoX metric that quantifies knowledge contributions. More details can be found in the Appendix.

% \begingroup
% \renewcommand\thefootnote{}
% \footnotetext{
% Our BiMind model and the tested datasets are accessible via the link:
% \href{https://github.com/cvzh/BiMind}{https://github.com/cvzh/BiMind}
% }
% \endgroup

\subsection{Problem Definitions}
% Information units usually involve most of the practical elements, such as sentiment, behavior, interaction, etc. Generally, these elements are represented by nouns, verbs, adjectives, etc., or hidden behind the words and sentences in the information content. In addition, these elements cover local semantic features and global context information. 

In this paper, we define the incorrect information detection as assessing whether a given information unit \(x_i\) is correct, where \(i\) is the \(i\)-th piece of information. Our detection foundation is that \(x_i\) is correct if no detected incorrectness exists. Therefore, the detection task is reframed as identifying incorrect elements within \(x_i\). Formally, we define:
\begin{equation}
    y(x_i) =
\begin{cases}
1 & \text{if  } \mathcal{I}(x_i) = \varnothing \quad \\
0 & \text{otherwise} \quad
\end{cases}
\end{equation}
Here, \(x_i\) represents title, sentence, article, or narrative. \(\mathcal{I}(x_i)\) is a set of incorrectness identified in \(x_i\), such as linguistic elements (tokens or phrases), representation elements (feature embeddings), or knowledge elements (retrieved neighbors). \(y(x_i)\) denotes correctness, i.e., 1 (correct information) or 0 (incorrect information).
For incorrect information detection, we model it as a binary classification function:
\begin{equation}
f(x_i) \rightarrow y(x_i)
\end{equation}
using a set of labeled training textual data, i.e.,
\begin{equation}
D_{\text{train}} = \left\{ (x_i, y(x_i)) \right\}_{i=1}^{|D_{\text{train}}|}
\end{equation}
\(y(x_i)\) is the label of \(x_i\). \(|D_{\text{train}}|\) is the total number of information units in the training dataset. We aim at learning the classification function:
\begin{equation}
f(x_i; \boldsymbol{\theta}) = \hat{y}_i
\end{equation}
where \(\hat{y}_i \in \{0, 1\}\) denotes the predicted label of $x_i$ and $\boldsymbol{\theta}$ is the learnable parameter vector.
% By minimizing the following objective function, our proposed model can predict the veracity of unseen content instances:
% \begin{align} as 
% \mathcal{L}(\boldsymbol{\theta}) = 
% & -\frac{1}{|D_{\text{train}}|} \sum_{i=1}^{|D_{\text{train}}|} \Big[ 
%       C(x_i) \log(\hat{y}_i) \notag \\
% & \quad + (1 - C(x_i)) \log(1 - \hat{y}_i) \Big]
% \end{align}

\begin{figure}
  \includegraphics[width=\columnwidth]{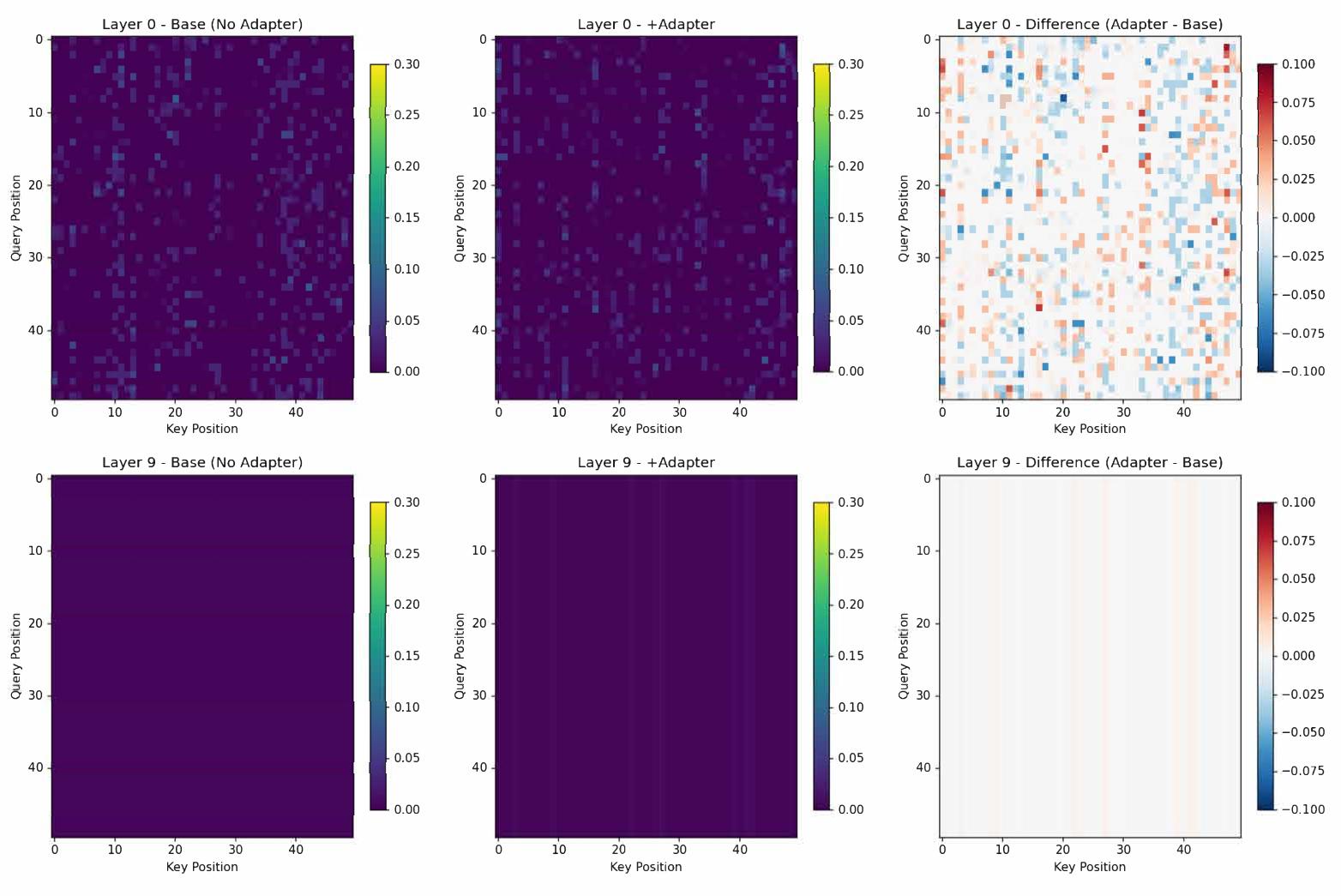}
  \caption{Illustration of the attention maps with ($+$) and without AGA across shallow (Layer 0) and deep (Layer 9) layers. AGA induces a structured, token-conditioned attention pattern at deeper layers, reshaping attention distribution by injecting global, key-focused inductive bias.}
  \label{fig:attenmap}
\end{figure}

\subsection{Attention Geometry Adapter}
Let $x_i = (t_1, t_2, \ldots, t_L)$ be a tokenized text sequence. $t_i$ denotes the
$i$-th token and $L$ is the length of the token sequence. Each token $t_i$ is mapped to an embedding $\boldsymbol{e}_i \in \mathbb{R}^{d}$ through an embedding matrix $\boldsymbol{E}$, where $d$ is the dimension of token embedding:
\begin{equation}
\boldsymbol{E}(x_i) = [\boldsymbol{e}_1, \boldsymbol{e}_2, \ldots, \boldsymbol{e}_L]
\in \mathbb{R}^{L \times d}
\end{equation}
Instead of directly modifying token embeddings, we introduce an AGA module that reshapes attention distributions at the pre-softmax logit level. It's a lightweight module that improves how the model focuses on tokens using grammar rules. For each token $t_i$, we assign a token-level attribute vector
$\boldsymbol{p}_i \in \{0,1\}^{D_{pos}}$ by using part-of-speech (POS) categories, where $D_{pos}$ is the number of POS tags. In our POS tag set, we set $D_{pos}=5$, including \textsc{VERB/AUX}, \textsc{NOUN}, \textsc{ADJ}, \textsc{ADV}, and \textsc{OTHER}.
This representation provides an interpretable, low-dimensional basis for conditioning attention geometry, where attention geometry denotes the structural pattern of attention distributions across tokens, beyond individual attention weights.

Then, the Transformer encoder projects token embeddings $\boldsymbol{E}(x_i)$ into queries, keys, and values as in the standard self-attention. For each attention head $h$, AGA computes token-conditioned logit offsets $\boldsymbol{\Delta}$ of query and key via lightweight multilayer perceptrons (MLPs):
\begin{equation}
\boldsymbol{\Delta}^{(\boldsymbol{q})} = f_q(\boldsymbol{p}_i), \qquad
\boldsymbol{\Delta}^{(\boldsymbol{k})} = f_k(\boldsymbol{p}_i)
\end{equation}
where $f_{q}(\cdot)$ and $f_{k}(\cdot)$ are two-layer MLPs. The final pre-softmax attention logits are updated as:
\begin{equation}
\widetilde{\boldsymbol{A}}^{(h)}_{i,j}
=
\frac{\boldsymbol{q}^{(h)\top}_i \boldsymbol{k}^{(h)}_j}{\sqrt{d_k}}
+
\boldsymbol{\Delta}^{(\boldsymbol{q})}_{h,i}
+
\boldsymbol{\Delta}^{(\boldsymbol{k})}_{h,j},
\end{equation}
where $\tilde{\boldsymbol{A}}^{(h)}_{i,j}$ denotes the pre-softmax attention logit between query token $\boldsymbol{q}^{(h)}_i$ and key token $\boldsymbol{k}^{(h)}_j$ in attention head $h$; $d_k$ is the dimension of the key vectors; $\boldsymbol{\Delta}^{(q)}_{h,i}$ and $\boldsymbol{\Delta}^{(k)}_{h,j}$ are token-conditioned offsets for head $h$ applied to the query and key logits, respectively. Through Eq.(7), AGA adjusts attention score by adding token-level biases before softmax. More details are provided in the Appendix. 

By injecting structured offsets, AGA reshapes the attention distributions, increasing entropy and mitigating attention collapse while leaving token embeddings unchanged, as shown in Figure~\ref{fig:attenmap}.
The attention outputs are then computed following standard multi-head attention
and passed through the subsequent layers of Transformer $\mathcal{T}_{AGA}(\cdot)$:
% \begin{equation}
% \boldsymbol{H} = \mathcal{T}(\boldsymbol{E}(x)) \in \mathbb{R}^{L \times d}.
% \end{equation}
\begin{equation}
   \boldsymbol{H}_i = \mathcal{T}_{\text{AGA}}(\boldsymbol{E}(x_i)) \in \mathbb{R}^{L\times d}
\end{equation}
in which $\boldsymbol{H}_i$ is the sequence representation. Finally, we apply max-pooling to capture the most salient features $\boldsymbol{h}$ in the sequence:
\begin{equation}
\boldsymbol{h} = \max_{i=1}^L \boldsymbol{H}_i \in \mathbb{R}^d
\end{equation}

\subsection{Semantic Neighbor Retrieval}
In this section, we construct an in-domain semantic memory $\boldsymbol{M} = \{\boldsymbol{m}_1, \boldsymbol{m}_2, \ldots, \boldsymbol{m}_N\}$ by encoding all training information units with a pre-trained SentenceTransformer \citep{reimers-2019-sentence-bert}, then feeding knowledge vectors into backbone models like LLaMA-7B \citep{touvron2023llama}. This memory serves as a lookup module that helps model find similar examples. $\boldsymbol{m}_j \in \mathbb{R}^{d_s}$ is the embedding of one information unit, normalized to unit length. $d_s$ is the dimension of knowledge embeddings, and $N$ is the number of training units stored in $\boldsymbol{M}$. For an input $x_i$, we encode it as query $\boldsymbol{q} \in \mathbb{R}^{d_s}$ and compute cosine similarity $s_i$ between $q$ and $m$:
\begin{equation}
s_j = \boldsymbol{q}^\top \boldsymbol{m}_j, \quad j=1,\ldots,N
\end{equation}
Next, we select the top-$k$ neighbors with indices $\mathcal{N}(x_i)$ and aggregate:
\begin{equation}
\bar{\boldsymbol{m}} = \frac{1}{k}\sum_{j \in \mathcal{N}(x_i)} \boldsymbol{m}_j.
\end{equation}
% We also compute similarity statistics to measure cross-modal agreement between heads, providing a geometric representation of consistency:
% \begin{equation}
% s_{\max} = \max_{j \in \mathcal{N}(x_i)} s_j, \quad s_{\text{mean}} = \frac{1}{k}\sum_{j \in \mathcal{N}(x_i)} s_j
% \end{equation}

To inject retrieved knowledge neighbors, we map $\bar{\boldsymbol{m}}$ into modulation parameters $\boldsymbol{\gamma}_k$ and $\boldsymbol{\beta}_k$:
\begin{equation}
\boldsymbol{\gamma}_k = \boldsymbol{W}_\gamma \bar{\boldsymbol{m}} + \boldsymbol{b}_\gamma, \quad \boldsymbol{\beta}_k = \boldsymbol{W}_\beta \bar{m} + \boldsymbol{b}_\beta
\end{equation}
Here, $\boldsymbol{W}_\gamma$, $\boldsymbol{W}_\beta$ $\in \mathbb{R}^{d \times d_s}$ are projection matrices; $\boldsymbol{b}_\gamma$ and $\boldsymbol{b}_\beta$ are bias terms. Then, we apply FiLM layers \citep{perez2018film} to produce the knowledge-augmented representation $\boldsymbol{h}_E$:
\begin{equation}
\boldsymbol{h}_E = \boldsymbol{h} \odot (1 + \tanh(\boldsymbol{\gamma}_k)) + \boldsymbol{\beta}_k
\end{equation}
where $\odot$ is element-wise multiplication, and FilM adaptively scales $(1 + \tanh(\boldsymbol{\gamma}_k))$ and shifts $(\boldsymbol{\beta}_k)$ each dimension of $\boldsymbol{h}$ based on knowledge retrieved from semantic memory. 
% , creating a knowledge-augmented representation $\boldsymbol{h}_E$
\subsection{Dual-head Prediction}
Combining text representations with content features $\boldsymbol{c} \in \mathbb{R}^{d_c}$, we define two reasoning heads:
\begin{equation}
\boldsymbol{z}_0 = \boldsymbol{W}_0 [\boldsymbol{h}; \boldsymbol{c}] + \boldsymbol{b}_0, \quad
\boldsymbol{z}_E = \boldsymbol{W}_E [\boldsymbol{h}_E; \boldsymbol{c}] + \boldsymbol{b}_E
\end{equation}
where $\boldsymbol{z}_0$ and $\boldsymbol{z}_E \in \mathbb{R}^K$ are the outputs of the content-internal and knowledge-augmented reasoning heads, respectively. $K$ is the number of labels. $\boldsymbol{W}_0$, $\boldsymbol{W}_E$ $\in \mathbb{R}^{d \times (d + d_c)}$ are projection matrices; $\boldsymbol{b}_0$ and $\boldsymbol{b}_E$ are bias terms. Then, $\boldsymbol{z}_0$, $\boldsymbol{z}_E$ are transformed into probability distributions:
\begin{equation}
\hat{\boldsymbol{y}}_0 = \text{softmax}(\boldsymbol{z}_0), \quad \hat{\boldsymbol{y}}_E = \text{softmax}(\boldsymbol{z}_E)
\end{equation}
$\hat{\boldsymbol{y}}_0$ is the prediction result from content-internal reasoning Head$_0$, and it's what model "believes" without external knowledge. Head$_E$ reflects knowledge-augmented reasoning and derives $\hat{\boldsymbol{y}}_E$, and $\hat{\boldsymbol{y}}_E$ is what model "believes" with knowledge. Together, we can explore how knowledge affects predictions.

\subsection{Uncertainty-aware Fusion Strategy}
We fuse two predictions from Head$_0$ and Head$_E$ using an entropy-gated strategy and a trainable agreement head, respectively. Using the entropy-gated fusion strategy, we first compute entropy for each head:
\begin{equation}
\mathcal{H}(\hat{\boldsymbol{y}}) = -\sum_{k=1}^K \hat{\boldsymbol{y}}_k \log \hat{\boldsymbol{y}}_k
\end{equation}
where $\hat{\boldsymbol{y}}$ is the predicted probability distribution produced by a reasoning head and $\hat{\boldsymbol{y}_k}$ denotes the probability of class $k$. Then, we formulate the gate input vector $\boldsymbol{u}$, and feed it into MLP gate:
% \begin{equation}
%     \boldsymbol{u} = [\boldsymbol{h}; \boldsymbol{h}_E; \mathcal{H}(\hat{\boldsymbol{y}}_0); \mathcal{H}(\hat{\boldsymbol{y}}_E)]
% \end{equation}
\begin{equation}
g = \sigma(\boldsymbol{W}_g \boldsymbol{u} + \boldsymbol{b}_g) \in (0,1) 
\end{equation}
$g$ is the fusion weight and $\sigma$ is the activation function. $\boldsymbol{W}_g$ and $\boldsymbol{b}_g$ are parameters of the MLP gate. Finally, we fuse logits:
\begin{equation}
\boldsymbol{z}_F = g \cdot \boldsymbol{z}_0 + (1-g) \cdot \boldsymbol{z}_E, \quad
\hat{\boldsymbol{y}}_F = \text{softmax}(\boldsymbol{z}_F) 
\end{equation}
where $\boldsymbol{z}_F$ is the fused logits combining content-internal and knowledge-augmented reasoning. $\hat{\boldsymbol{y}}_F$ is the final prediction results. Therefore, if Head$_0$ has high entropy (uncertainty), the gate shifts the weight toward Head$_E$, and vice versa. More details can be found in the Appendix.

\subsection{Agreement Regularization}
To stabilize training process, we enforce the agreement between heads while preserving differences. We define the symmetric KL regularizer as:
\begin{equation}
\mathcal{L}^{(i)}_{\text{agree}} = \tfrac{1}{2}\left[D_{\text{KL}}(\hat{\boldsymbol{y}}^{(i)}_0 \,\|\, \hat{\boldsymbol{y}}^{(i)}_E) + D_{\text{KL}}(\hat{\boldsymbol{y}}^{(i)}_E \,\|\, \hat{\boldsymbol{y}}^{(i)}_0)\right]
\end{equation}
where $\mathcal{L}^{(i)}_{\text{agree}}$ is agreement loss. $D_{\text{KL}}(\hat{\boldsymbol{y}}^{(i)}_0 \,\|\, \hat{\boldsymbol{y}}^{(i)}_E)$ is the KL divergence between distributions $\hat{\boldsymbol{y}}^{(i)}_0$ and $\hat{\boldsymbol{y}}^{(i)}_E$. Based on the symmetric KL regularizer, our training objective is:
\begin{align}
\mathcal{L} = \frac{1}{n} \sum_{i=1}^n \Big[ &
\mathcal{L}_{\text{CE}}(\boldsymbol{z}^{(i)}_F, y(x_i)) 
+ \tfrac{1}{2}\,\mathcal{L}_{\text{CE}}(\boldsymbol{z}^{(i)}_0, y(x_i)) \nonumber \\
&+ \tfrac{1}{2}\,\mathcal{L}_{\text{CE}}(\boldsymbol{z}^{(i)}_E, y(x_i)) 
+ \lambda\,\mathcal{L}_{\text{agree}}^{(i)} \Big]
\end{align}
where $\mathcal{L}^{(i)}_{\text{CE}}$ is cross-entropy (CE) loss, and $\lambda$ is the agreement loss weight. Through loss function $\mathcal{L}$, we encourage both heads to produce consistent predictions while maintaining their distinct reasoning.

\subsection{Value-of-eXperience Metric}
To measure knowledge contributions, we define VoX at the instance level. Terminologically, we refer to the retrieved knowledge as "experience" in our framework, to clarify its role as external evidence augmenting content-internal predictions. Given the correctness label $y(x_i)$, VoX is:
\begin{equation}
\text{VoX}(x_i) = \boldsymbol{z}_E[y(x_i)] - \boldsymbol{z}_0[y(x_i)].
\end{equation}
Our interpretations are summarized as follows:
\begin{itemize}
    \item $\text{VoX}(x_i) > 0$: knowledge increases prediction confidence in correct class.
    \item $\text{VoX}(x_i) < 0$: knowledge decreases prediction confidence, suggesting potential noise.
    \item $\text{VoX}(x_i) \approx 0$: knowledge has little effect.
\end{itemize}
%We report mean, median, and the proportion of positive-VoX samples to provide interpretable diagnostics.
Unlike accuracy or F1 score, VoX highlights when and why knowledge matters and provides interpretable diagnostics on knowledge augmentation.

\section{Experiments}
In this section, we conducted extensive experiments on four datasets collected from real-world scenarios, and experimental results demonstrate that our model has superior performance and efficiency to most tested models. We first introduced the experimental setup, including the datasets, tested models, and experimental settings. Then, we reported the experiment results and VoX values, and then analyzed these results for further exploration. Furthermore, the ablation study shows the modules contributing to the performance improvement. More details are provided in the Appendix.
% In addition, we analyze the key parameters to validate our model's sensitivity to these parameters. 

\subsection{Experimental Setup}
\textbf{Datasets}. For conducting extensive experiments, we used four datasets to broadly test our model and other advanced models, including health datasets (MM COVID \citep{Li2020MMCOVIDAM} and RoCOVery \citep{Zhou2020ReCOVeryAM}), news content dataset (LIAR \citep{Wang2017LiarLP}), and a multi-domain dataset (MC Fake \cite{Min2022DivideandConquerPI}).

\begin{table*}[t]
\scriptsize
\centering
\setlength{\tabcolsep}{1pt}
\renewcommand{\arraystretch}{1.0}
\resizebox{\linewidth}{!}{
\begin{tabular}{l|cc|cc|cc|cc|cc|cc}
% \rowcolor{lightorange}
\multicolumn{13}{c}{} \\
\toprule
\textbf{Dataset} 
& \multicolumn{2}{c|}{\textbf{CNN}} 
& \multicolumn{2}{c|}{\textbf{GCN}} 
& \multicolumn{2}{c|}{\textbf{BERT}} 
& \multicolumn{2}{c|}{\textbf{HAN}} 
& \multicolumn{2}{c|}{\textbf{HeteroSGT}} 
& \multicolumn{2}{c}{\textbf{BiMind}} \\
\cmidrule{2-13}
& Acc & Pre & Acc & Pre & Acc & Pre & Acc & Pre & Acc & Pre & Acc & Pre \\
\midrule
\rowcolor{rowgray}
MM COVID & 0.582±0.035 & 0.478±0.170 & 0.717±0.156 & 0.735±0.236 & 0.730±0.093 & 0.727±0.094 & {0.855±0.005} & {0.854±0.005} & \textcolor{blue}{0.915±0.009} & \textcolor{blue}{0.905±0.011} & \textcolor{red}{0.951±0.008} & \textcolor{red}{0.950±0.011} \\
ReCOVery & 0.658±0.011 & 0.460±0.104 & 0.718±0.037 & 0.691±0.178 & 0.682±0.030 & 0.441±0.213 & 0.722±0.021 & 0.462±0.197 & \textcolor{blue}{0.727±0.023} & \textcolor{blue}{0.731±0.047} & \textcolor{red}{0.918±0.013} & \textcolor{red}{0.922±0.013} \\
\rowcolor{rowgray}
MC Fake & 0.825±0.001 & 0.544±0.156 & 0.724±0.138 & 0.516±0.169 & {0.827±0.006} & {0.713±0.271} & 0.825±0.005 & 0.463±0.098 & \textcolor{blue}{0.883±0.002} & \textcolor{blue}{0.812±0.003} & \textcolor{red}{0.887±0.005} & \textcolor{red}{0.827±0.006} \\
LIAR     & 0.546±0.019 & 0.432±0.181 & 0.487±0.039 & 0.493±0.047 & 0.537±0.007 & 0.513±0.017 & 0.546±0.025 & 0.493±0.036 & \textcolor{blue}{0.581±0.002} & \textcolor{blue}{0.580±0.003} & \textcolor{red}{0.633±0.001} & \textcolor{red}{0.637±0.002} \\
\midrule
\textbf{Dataset} 
& Rec & F1 & Rec & F1 & Rec & F1 & Rec & F1 & Rec & F1 & Rec & F1 \\
\midrule
\rowcolor{rowgray}
MM COVID & 0.547±0.039 & 0.474±0.101 & 0.685±0.178 & 0.621±0.184 & 0.722±0.101 & 0.720±0.103 & {0.854±0.006} & {0.853±0.005} & \textcolor{blue}{0.883±0.013} & \textcolor{blue}{0.893±0.011} & \textcolor{red}{0.951±0.009} & \textcolor{red}{0.951±0.008} \\
ReCOVery & 0.501±0.020 & 0.422±0.107 & 0.609±0.102 & 0.516±0.021 & 0.722±0.081 & 0.416±0.032 & 0.506±0.002 & 0.457±0.013 & \textcolor{blue}{0.585±0.036} & \textcolor{blue}{0.571±0.049}  & \textcolor{red}{0.918±0.013} & \textcolor{red}{0.919±0.013} \\
\rowcolor{rowgray}
MC Fake  & 0.501±0.002 & 0.455±0.004 & {0.552±0.169} & {0.470±0.039} & 0.502±0.001 & 0.451±0.002 & 0.500±0.004 & 0.453±0.001 & \textcolor{red}{0.762±0.002} & \textcolor{red}{0.783±0.003} & \textcolor{blue}{0.700±0.099} & \textcolor{blue}{0.738±0.109} \\
LIAR     & 0.502±0.005 & 0.377±0.049 & 0.494±0.029 & 0.423±0.055 & 0.510±0.012 & 0.483±0.014 & 0.502±0.018 & 0.445±0.053 & \textcolor{blue}{0.575±0.002} & \textcolor{blue}{0.571±0.003} & \textcolor{red}{0.636±0.003} & \textcolor{red}{0.633±0.002} \\
\bottomrule
\end{tabular}
}
\caption{Detection performance on four datasets (best in \textcolor{red}{red}, second-best in \textcolor{blue}{blue}).}
\label{tab:classification-performance}
\end{table*}

\textbf{Experimental Models}. To fairly perform the comparison experiments, we compared our proposed BiMind model with five models, which include a CNN-based model \citep{Kim2014ConvolutionalNN}, a GCN-based model \citep{Yao2018GraphCN}, HAN \citep{Yang2016HierarchicalAN}, BERT \citep{devlin2019bert}, and HeteroSGT \citep{Zhang2024HeterogeneousST}.

\subsection{Experiment Settings}
For training and testing our proposed model, we split all the datasets into train, validation, and test datasets using a ratio of 80\%, 10\%, and 10\%, respectively. To validate the generalizability of tested methods, we performed 10 rounds of tests with random seeds for each model and then recorded the average results and standard deviation. Here, all the experiments were conducted on 1 NVIDIA A100 GPU with 40 GB RAM. We quantitatively evaluated our model's performance compared to the other five tested models, using classification metrics such as accuracy (Acc), Macro-precision (Pre), Macro-F1 (F1), and Macro-recall (Rec).

\subsection{Experimental Results}
In Table~\ref{tab:classification-performance}, we reported the experimental results of all the tested models across the four datasets. From Table~\ref{tab:classification-performance}, one can see that our model achieves superior performance across all the metrics on the MM COVID, LIAR, and ReCOVery datasets, and suboptimal performance on the dataset MC Fake. It shows that our modules can improve the model performance and have a significant impact on detecting incorrect information. Additionally, we can see that our model achieves higher recall values on all four datasets, typically on the MM COVID, LIAR, and ReCOVery datasets. A higher recall value indicates that less incorrect information is missed. Furthermore, it should be noted that our model has robust and consistent performance across all the datasets, compared with other tested models. Though HeteroSGT achieves the optimal results, such as Rec and F1 on MC Fake due to its subgraph structure, it still drops performance by 19.1$\%$ on Acc and 19.1$\%$ on Pre, 5.2$\%$ on Acc and 5.7$\%$ on Pre, respectively, compared to our proposed model on ReCOVery and LIAR datasets. More details are provided in the Appendix.

\begin{table*}[t]
\centering
\setlength{\tabcolsep}{15pt}
\resizebox{2.0\columnwidth}{!}{
\begin{tabular}{l l r r r r l l}
\toprule
\textbf{Dataset} & \textbf{Head/Mode} & \textbf{Acc} & \textbf{F1} & \textbf{Pre} & \textbf{Rec} & \textbf{VoX (mean / pos\%)} & \textbf{Gate Mean (\%$<$0.3 / \%$>$0.7)} \\
\midrule
\multirow{3}{*}{ReCOVery}
  & $\text{Head}_0$    & 85.22 & 85.15 & 85.10 & 85.22 & -- & -- \\
  & $\text{Head}_E$  & 87.19 & 86.93 & 86.98 & 87.19 & +0.47 / 84.24\% & 0.04 (100.00\% / 0.00\%) \\
  & Fused     & 87.19 & 86.93 & 86.98 & 87.19 & -- & 0.04 (100.00\% / 0.00\%) \\
\midrule
\multirow{3}{*}{MC Fake}
  & $\text{Head}_0$    & 86.50 & 84.85 & 84.79 & 86.50 & -- & -- \\
  & $\text{Head}_E$  & 87.35 & 86.90 & 86.63 & 87.35 & -0.08 / 39.91\% & 0.22 (81.24\% / 0.00\%) \\
  & Fused     & 87.42 & 86.82 & 86.52 & 87.42 & -- & 0.22 (81.24\% / 0.00\%) \\
\midrule
\multirow{3}{*}{MM COVID}
  & $\text{Head}_0$    & 81.70 & 81.78 & 82.13 & 81.70 & -- & -- \\
  & $\text{Head}_E$  & 90.45 & 90.51 & 91.18 & 90.45 & +0.97 / 83.29\% & 0.03 (100.00\% / 0.00\%) \\
  & Fused     & 90.72 & 90.77 & 91.38 & 90.72 & -- & 0.03 (100.00\% / 0.00\%) \\
\midrule
\multirow{3}{*}{LIAR}
  & $\text{Head}_0$   & 59.90 & 58.78 & 59.66 & 59.90 & -- & -- \\
  & $\text{Head}_E$  & 57.70 & 47.30 & 66.32 & 57.70 & +0.07 / 60.56\% & 0.19 (100.00\% / 0.00\%) \\
  & Fused     & 58.29 & 48.66 & 66.51 & 58.29 & -- & 0.19 (100.00\% / 0.00\%) \\
\bottomrule
\end{tabular}
}
\caption{Cross-dataset VoX results. We report mean VoX value and percentage of samples with positive gain (pos\%); we also show mean gate value and routing mass below/above thresholds, with \%$<$0.3 indicating the gate leaned strongly toward knowledge head $\text{Head}_E$, and \%$>$0.7 indicating it leaned strongly toward content head $\text{Head}_0$.}
\label{tab:bimind_cross_dataset}
\end{table*}

\begin{table}[h]
\centering
\scriptsize
\setlength{\tabcolsep}{5pt}
\renewcommand{\arraystretch}{1.2}
\begin{tabular}{l|c|c|c|c}
\toprule
\textbf{Model Variant} & \textbf{Acc} & \textbf{Pre} & \textbf{Rec} & \textbf{F1} \\
\midrule
\textbf{Full BiMind model (w/ knowledge)} 
& \textbf{0.897} & \textbf{0.895} & \textbf{0.897} & \textbf{0.895} \\
Baseline (content w/o knowledge)      
& 0.852 & 0.849 & 0.852 & 0.848 \\
\midrule
-- Attention geometry adapter
& 0.872 & 0.870 & 0.872 & 0.870 \\
-- Knowledge retrieval 
& 0.847 & 0.847 & 0.847 & 0.847 \\
-- Gated fusion  
& 0.862 & 0.861 & 0.862 & 0.861 \\
-- Trainable agreement head
& 0.867 & 0.868 & 0.867 & 0.867 \\
-- Symmetric KL Regularizer
& 0.872 & 0.881 & 0.872 & 0.874 \\
\bottomrule
\end{tabular}
\caption{Ablation study on ReCOVery dataset. One module is removed at a time from the full BiMind model.}
\label{tab:ablation}
\end{table}

% \subsection{VoX Analysis}
\subsection{Case Study}
As illustrated in Table~\ref{tab:bimind_cross_dataset}, the performance of BiMind varies across benchmark datasets due to the gate routes between Head$_0$ and Head$_E$. On datasets ReCOVery ($+$0.47 / 84.24\% VoX, $\approx$0.04 Gate) and MM COVID ($+$0.97 / 83.29\% VoX, $\approx$0.03 Gate), the gate leans strongly towards Head$_E$ because the retrieved knowledge positively aligned with ground-truth labels. On MC Fake ($-$0.08 / 39.91\% VoX, $\approx$0.22 Gate), the fusion has partial reliance on knowledge and generates mixed results, which improve minority-class recall but introduce noise. In contrast, on LIAR ($+$0.07 / 60.56\% VoX, $\approx$0.19 Gate), we can see that when external knowledge is noisy, it leads fusion to weaken the predictions with a significant drop in F1 (58.78 $\rightarrow$ 47.30), highlighting that low gate values are not always effective and thus must be interpreted in the context of knowledge integrity and veracity. Additional analysis can be found in the Appendix.

\subsection{Ablation Study}
We conducted an ablation study on the ReCOVery dataset to evaluate the performance of four core modules in our model: AGA, self-retrieved knowledge module via FiLM, fusion strategies (entropy-gated scheme and trainable agreement head), and symmetric KL regularizer. From Table~\ref{tab:ablation}, we can see that the BiMind model, with all these four modules, achieved the best performance, i.e., Acc of 0.897, Pre of 0.895, Rec of 0.897, and F1 score of 0.895. More results can be found in the Appendix.

\section{Limitations}
Though our proposed BiMind framework has superior performance in the incorrect information detection task by integrating textual and knowledge features, several limitations remain. First, AGA conditions attention geometry on token-level attributes, which might be less efficient for inputs with limited salient lexical signals. Secondly, BiMind does not incorporate social credibility or propagation patterns into the detection pipeline. Then, when the detection model has prediction errors, it might weaken the correct information flow. Finally, with the growing prevalence of multimodal information on social media platforms, incorrect information is not limited to textual content but involves images, videos, and audio. Therefore, multimodal incorrect information detection has become a critical challenge. BiMind can be extended to a multimodal reasoning framework by the disentanglement principle: one head focuses on content-internal textual reasoning, and another head encodes visual signal, followed by uncertainty-aware fusion to calibrate their contributions.

\section{Conclusion}
Incorrect information significantly disrupts content quality and integrity on social media platforms, and therefore, it's increasingly important to develop detection models that are efficient and interpretable. Compared to most detection approaches that blend textual content and external knowledge, we proposed \textbf{BiMind}, a dual-head framework that explicitly disentangles \emph{content-internal reasoning} from \emph{knowledge-augmented reasoning} for incorrect information detection. In this work, we first designed an attention geometry adapter that reshapes attention distributions to prevent attention collapse. Secondly, an in-memory semantic knowledge base was constructed to retrieve and encode external knowledge features through the FiLM layer. Then, we introduced two uncertainty-aware fusion strategies, including an entropy-gated strategy and a trainable agreement head, regularized by a symmetric KL regularizer. Finally, the VoX metric was defined, which quantifies the knowledge contributions, providing interpretable diagnostics at the instance level on \emph{when} and \emph{why} knowledge impacts detection. Experimental results across benchmark datasets show that BiMind achieves competitive performance while providing interpretable insights on knowledge injections.

\section*{Acknowledgments}
This work is supported in part by the National Science Foundation under Award No. 2343387.

% This document has been adapted
% by Steven Bethard, Ryan Cotterell and Rui Yan
% from the instructions for earlier ACL and NAACL proceedings, including those for
% ACL 2019 by Douwe Kiela and Ivan Vuli\'{c},
% NAACL 2019 by Stephanie Lukin and Alla Roskovskaya,
% ACL 2018 by Shay Cohen, Kevin Gimpel, and Wei Lu,
% NAACL 2018 by Margaret Mitchell and Stephanie Lukin,
% Bib\TeX{} suggestions for (NA)ACL 2017/2018 from Jason Eisner,
% ACL 2017 by Dan Gildea and Min-Yen Kan,
% NAACL 2017 by Margaret Mitchell,
% ACL 2012 by Maggie Li and Michael White,
% ACL 2010 by Jing-Shin Chang and Philipp Koehn,
% ACL 2008 by Johanna D. Moore, Simone Teufel, James Allan, and Sadaoki Furui,
% ACL 2005 by Hwee Tou Ng and Kemal Oflazer,
% ACL 2002 by Eugene Charniak and Dekang Lin,
% and earlier ACL and EACL formats written by several people, including
% John Chen, Henry S. Thompson and Donald Walker.
% Additional elements were taken from the formatting instructions of the \emph{International Joint Conference on Artificial Intelligence} and the \emph{Conference on Computer Vision and Pattern Recognition}.

\bibliography{custom}

@String{Computing = "Computing" }

@String{Springer = "Springer-Verlag" }

@ArtifactSoftware{R,
    title = {R: A Language and Environment for Statistical Computing},
    author = {{R Core Team}},
    organization = {R Foundation for Statistical Computing},
    address = {Vienna, Austria},
    year = {2019},
    url = {https://www.R-project.org/},
}

@article{shu2017fake,
author = {Shu, Kai and Sliva, Amy and Wang, Suhang and Tang, Jiliang and Liu, Huan},
title = {Fake news detection on social media: a data mining perspective},
year = {2017},
issue_date = {June 2017},
publisher = {Association for Computing Machinery},
address = {New York, NY, USA},
volume = {19},
number = {1},
issn = {1931-0145},
url = {https://doi.org/10.1145/3137597.3137600},
doi = {10.1145/3137597.3137600},
journal = {SIGKDD Explor. Newsl.},
month = sep,
pages = {22–36},
numpages = {15}
}

@article{zhou2020survey,
author = {Zhou, Xinyi and Zafarani, Reza},
title = {A Survey of Fake News: Fundamental Theories, Detection Methods, and Opportunities},
year = {2020},
issue_date = {September 2021},
publisher = {Association for Computing Machinery},
address = {New York, NY, USA},
volume = {53},
number = {5},
issn = {0360-0300},
url = {https://doi.org/10.1145/3395046},
doi = {10.1145/3395046},
journal = {ACM Computing Survery},
month = sep,
articleno = {109},
numpages = {40}
}

@article{ahmed2022combining,
  title={Combining machine learning with knowledge engineering to detect fake news in social networks-a survey},
  author={Ahmed, Sajjad and Hinkelmann, Knut and Corradini, Flavio},
  journal={arXiv preprint arXiv:2201.08032},
  year={2022},
  url={https://arxiv.org/abs/2201.08032}
}

@article{chen2026actormindemulatinghumanactor,
      title={ActorMind: Emulating Human Actor Reasoning for Speech Role-Playing}, 
      author={Chen, Xi and Xue, Wei and Guo, Yike},
      journal={arXiv preprint arXiv:2604.11103},
      year={2026},
      url={https://arxiv.org/abs/2604.11103}, 
}

@article{lan2025contextual,
  title={Contextual integrity in LLMs via reasoning and reinforcement learning},
  author={Lan, Guangchen and Inan, Huseyin A and Abdelnabi, Sahar and Kulkarni, Janardhan and Wutschitz, Lukas and Shokri, Reza and Brinton, Christopher G and Sim, Robert},
  journal={arXiv preprint arXiv:2506.04245},
  url={https://arxiv.org/abs/2506.04245}, 
  year={2025}
}

@article{bhattarai2021explainable,
  title={Explainable tsetlin machine framework for fake news detection with credibility score assessment},
  author={Bhattarai, Bimal and Granmo, Ole-Christoffer and Jiao, Lei},
  journal={arXiv preprint arXiv:2105.09114},
  year={2021},
  url={https://arxiv.org/abs/2105.09114}
}

@article{mazzeo2021detection,
  title={Detection of fake news on COVID-19 on web search engines},
  author={Mazzeo, Valeria and Rapisarda, Andrea and Giuffrida, Giovanni},
  journal={Frontiers in Physics},
  volume={9},
  pages={685730},
  year={2021},
  publisher={Frontiers Media SA},
  url={https://doi.org/10.3389/fphy.2021.685730}
}

@inproceedings{vlachos2014fact,
  title={Fact checking: Task definition and dataset construction},
  author={Vlachos, Andreas and Riedel, Sebastian},
  booktitle={Proceedings of the ACL 2014 Workshop on Language Technologies and Computational Social Science},
  pages={18--22},
  year={2014},
  url={https://aclanthology.org/W14-2508.pdf}
}

@inproceedings{hassan2015detecting,
author = {Hassan, Naeemul and Li, Chengkai and Tremayne, Mark},
title = {Detecting Check-worthy Factual Claims in Presidential Debates},
year = {2015},
isbn = {9781450337946},
url = {https://doi.org/10.1145/2806416.2806652},
doi = {10.1145/2806416.2806652},
booktitle = {Proceedings of the 24th ACM International on Conference on Information and Knowledge Management},
pages = {1835--1838}
}

@article{etzioni2008open,
  title={Open information extraction from the web},
  author={Etzioni, Oren and Banko, Michele and Soderland, Stephen and Weld, Daniel S},
  journal={Communications of the ACM},
  volume={51},
  number={12},
  pages={68--74},
  year={2008},
  publisher={ACM New York, NY, USA},
  url={https://dl.acm.org/doi/fullHtml/10.1145/1409360.1409378}
}

@inproceedings{wu2014toward,
author = {Wu, You and Agarwal, Pankaj K. and Li, Chengkai and Yang, Jun and Yu, Cong},
title = {Toward computational fact-checking},
year = {2014},
issue_date = {March 2014},
booktitle = {Proceedings of the VLDB Endowment},
issn = {2150-8097},
url = {https://doi.org/10.14778/2732286.2732295},
doi = {10.14778/2732286.2732295},
journal = {Proc. VLDB Endow.},
month = mar,
pages = {589--600},
numpages = {12}
}

@inproceedings{shi2016fact,
author = {Shi, Baoxu and Weninger, Tim},
title = {Fact Checking in heterogeneous information networks},
year = {2016},
isbn = {9781450341448},
url = {https://doi.org/10.1145/2872518.2889354},
doi = {10.1145/2872518.2889354},
booktitle = {Proceedings of the 25th International Conference Companion on World Wide Web},
pages = {101--102}
}

@article{su2020motivations,
  title={Motivations, methods and metrics of misinformation detection: an NLP perspective},
  author={Su, Qi and Wan, Mingyu and Liu, Xiaoqian and Huang, Chu-Ren},
  journal={Natural Language Processing Research},
  volume={1},
  number={1},
  pages={1--13},
  year={2020},
  publisher={Springer},
  url={https://doi.org/10.2991/nlpr.d.200522.001}
}

@inproceedings{ruchansky2017csi,
author = {Ruchansky, Natali and Seo, Sungyong and Liu, Yan},
title = {CSI: A Hybrid Deep Model for Fake News Detection},
year = {2017},
isbn = {9781450349185},
url = {https://doi.org/10.1145/3132847.3132877},
doi = {10.1145/3132847.3132877},
booktitle = {Proceedings of the 2017 ACM on Conference on Information and Knowledge Management},
pages = {797--806}
}

@article{kaliyar2020fndnet,
  title={FNDNet--a deep convolutional neural network for fake news detection},
  author={Kaliyar, Rohit Kumar and Goswami, Anurag and Narang, Pratik and Sinha, Soumendu},
  journal={Cognitive Systems Research},
  volume={61},
  pages={32--44},
  year={2020},
  publisher={Elsevier},
  url={https://doi.org/10.1016/j.cogsys.2019.12.005}
}

@article{kaliyar2021fakebert,
  title={FakeBERT: Fake news detection in social media with a BERT-based deep learning approach},
  author={Kaliyar, Rohit Kumar and Goswami, Anurag and Narang, Pratik},
  journal={Multimedia Tools and Applications},
  volume={80},
  number={8},
  pages={11765--11788},
  year={2021},
  publisher={Springer},
  url={https://doi.org/10.1007/s11042-020-10183-2}
}

@inproceedings{devlin2019bert,
  title={Bert: Pre-training of deep bidirectional transformers for language understanding},
  author={Devlin, Jacob and Chang, Ming-Wei and Lee, Kenton and Toutanova, Kristina},
  booktitle={Proceedings of the 2019 Conference of the North American Chapter of the Association for Computational Linguistics: Human Language Technologies, volume 1 (long and short papers)},
  pages={4171--4186},
  year={2019},
  url={https://aclanthology.org/N19-1423/}
}

@article{abdali2024multi,
  title={Multi-modal misinformation detection: Approaches, challenges and opportunities},
  author={Abdali, Sara and Shaham, Sina and Krishnamachari, Bhaskar},
  journal={ACM Computing Surveys},
  volume={57},
  number={3},
  pages={1--29},
  year={2024},
  publisher={ACM New York, NY},
  url={https://dl.acm.org/doi/full/10.1145/3697349}
}

@inproceedings{conneau2019cross,
  title={Cross-lingual language model pretraining},
  author={Conneau, Alexis and Lample, Guillaume},
  booktitle={Proceedings of the 33rd International Conference on Neural Information Processing Systems},
  pages={7059--7069},
  year={2019},
  url={https://dl.acm.org/doi/abs/10.5555/3454287.3454921}
}

@inproceedings{Shu2017BeyondNC,
author = {Shu, Kai and Wang, Suhang and Liu, Huan},
title = {Beyond News Contents: The Role of Social Context for Fake News Detection},
year = {2019},
isbn = {9781450359405},
url = {https://doi.org/10.1145/3289600.3290994},
doi = {10.1145/3289600.3290994},
booktitle = {Proceedings of the 12nd ACM International Conference on Web Search and Data Mining},
pages = {312--320}
}

@article{Zhou2019NetworkbasedFN,
  title={Network-based Fake News Detection: A Pattern-driven Approach},
  author={Xinyi Zhou and Reza Zafarani},
  journal={arXiv preprint arXiv:1906.04210},
  year={2019},
  url={https://api.semanticscholar.org/CorpusID:184487035}
}

@article{Fu2022DISCOCA,
  title={DISCO: Comprehensive and Explainable Disinformation Detection},
  author={Dongqi Fu and Yikun Ban and Hanghang Tong and Ross Maciejewski and Jingrui He},
  journal={Proceedings of the 31st ACM International Conference on Information \& Knowledge Management},
  year={2022},
  pages={4848--4852},
  url={https://api.semanticscholar.org/CorpusID:247318419}
}

@inproceedings{Zhu2024PropagationSG,
author = {Zhu, Junyou and Gao, Chao and Yin, Ze and Li, Xianghua and Kurths, Juergen},
title = {Propagation Structure-Aware Graph Transformer for Robust and Interpretable Fake News Detection},
year = {2024},
isbn = {9798400704901},
url = {https://doi.org/10.1145/3637528.3672024},
doi = {10.1145/3637528.3672024},
booktitle = {Proceedings of the 30th ACM SIGKDD Conference on Knowledge Discovery and Data Mining},
pages = {4652--4663}
}

@inproceedings{reimers-2019-sentence-bert,
    title = "Sentence-{BERT}: Sentence Embeddings using {S}iamese {BERT}-Networks",
    author = "Reimers, Nils  and
      Gurevych, Iryna",
    editor = "Inui, Kentaro  and
      Jiang, Jing  and
      Ng, Vincent  and
      Wan, Xiaojun",
    booktitle = "Proceedings of the 2019 Conference on Empirical Methods in Natural Language Processing and the 9th International Joint Conference on Natural Language Processing",
    month = nov,
    year = "2019",
    address = "Hong Kong, China",
    publisher = "Association for Computational Linguistics",
    url = "https://aclanthology.org/D19-1410/",
    doi = "10.18653/v1/D19-1410",
    pages = "3982--3992"
}

@article{Li2020MMCOVIDAM,
  title={MM-COVID: A Multilingual and Multimodal Data Repository for Combating COVID-19 Disinformation},
  author={Yichuan Li and Bohan Jiang and Kai Shu and Huan Liu},
  journal={arXiv preprint arXiv:2011.04088},
  year={2020},
  url={https://api.semanticscholar.org/CorpusID:227143124}
}

@inproceedings{Zhou2020ReCOVeryAM,
author = {Zhou, Xinyi and Mulay, Apurva and Ferrara, Emilio and Zafarani, Reza},
title = {ReCOVery: A Multimodal Repository for COVID-19 News Credibility Research},
year = {2020},
isbn = {9781450368599},
url = {https://doi.org/10.1145/3340531.3412880},
doi = {10.1145/3340531.3412880},
booktitle = {Proceedings of the 29th ACM International Conference on Information \& Knowledge Management},
pages = {3205--3212},
numpages = {8}
}

@inproceedings{Min2022DivideandConquerPI,
author = {Min, Erxue and Rong, Yu and Bian, Yatao and Xu, Tingyang and Zhao, Peilin and Huang, Junzhou and Ananiadou, Sophia},
title = {Divide-and-Conquer: Post-User Interaction Network for Fake News Detection on Social Media},
year = {2022},
isbn = {9781450390965},
url = {https://doi.org/10.1145/3485447.3512163},
doi = {10.1145/3485447.3512163},
booktitle = {Proceedings of the ACM Web Conference},
pages = {1148--1158},
}

@article{Yao2018GraphCN,
  title={Graph Convolutional Networks for Text Classification},
  author={Liang Yao and Chengsheng Mao and Yuan Luo},
  journal={arXiv preprint arXiv:1809.05679},
  year={2018},
  url={https://api.semanticscholar.org/CorpusID:52284222}
}

@inproceedings{Yang2016HierarchicalAN,
    title = "Hierarchical Attention Networks for Document Classification",
    author = "Yang, Zichao  and
      Yang, Diyi  and
      Dyer, Chris  and
      He, Xiaodong  and
      Smola, Alex  and
      Hovy, Eduard",
    editor = "Knight, Kevin  and
      Nenkova, Ani  and
      Rambow, Owen",
    booktitle = "Proceedings of the 2016 Conference of the North {A}merican Chapter of the Association for Computational Linguistics: Human Language Technologies",
    month = jun,
    year = "2016",
    address = "San Diego, California",
    publisher = "Association for Computational Linguistics",
    url = "https://aclanthology.org/N16-1174/",
    doi = "10.18653/v1/N16-1174",
    pages = "1480--1489"
}

@inproceedings{Zhang2024HeterogeneousST,
author = {Zhang, Yuchen and Ma, Xiaoxiao and Wu, Jia and Yang, Jian and Fan, Hao},
title = {Heterogeneous Subgraph Transformer for Fake News Detection},
year = {2024},
isbn = {9798400701719},
url = {https://doi.org/10.1145/3589334.3645680},
doi = {10.1145/3589334.3645680},
booktitle = {Proceedings of the ACM Web Conference},
pages = {1272--1282}
}

@article{tsao2021social,
  title={What social media told us in the time of COVID-19: a scoping review},
  author={Tsao, Shu-Feng and Chen, Helen and Tisseverasinghe, Therese and Yang, Yang and Li, Lianghua and Butt, Zahid A},
  journal={The Lancet Digital Health},
  volume={3},
  number={3},
  pages={e175--e194},
  year={2021},
  publisher={Elsevier},
  url={https:doi/10.1016/S2589-7500(20)30315-0}
}

@inproceedings{yang-etal-2023-rumor,
    title = "Rumor Detection on Social Media with Crowd Intelligence and {C}hat{GPT}-Assisted Networks",
    author = "Yang, Chang  and
      Zhang, Peng  and
      Qiao, Wenbo  and
      Gao, Hui  and
      Zhao, Jiaming",
    editor = "Bouamor, Houda  and
      Pino, Juan  and
      Bali, Kalika",
    booktitle = "Proceedings of the 2023 Conference on Empirical Methods in Natural Language Processing",
    month = dec,
    year = "2023",
    address = "Singapore",
    publisher = "Association for Computational Linguistics",
    url = "https://aclanthology.org/2023.emnlp-main.347/",
    doi = "10.18653/v1/2023.emnlp-main.347",
    pages = "5705--5717"
}

@article{10.1145/3393880,
author = {Guo, Bin and Ding, Yasan and Yao, Lina and Liang, Yunji and Yu, Zhiwen},
title = {The Future of False Information Detection on Social Media: New Perspectives and Trends},
year = {2020},
issue_date = {July 2021},
publisher = {Association for Computing Machinery},
address = {New York, NY, USA},
volume = {53},
number = {4},
issn = {0360-0300},
url = {https://doi.org/10.1145/3393880},
doi = {10.1145/3393880},
journal = {ACM Computing Survery},
pages = {1--36},
month = jul,
articleno = {68},
numpages = {36}
}

@inproceedings{shi-etal-2023-multiview,
    title = "Multiview Clickbait Detection via Jointly Modeling Subjective and Objective Preference",
    author = "Shi, Chongyang  and
      Yin, Yijun  and
      Zhang, Qi  and
      Xiao, Liang  and
      Naseem, Usman  and
      Wang, Shoujin  and
      Hu, Liang",
    editor = "Bouamor, Houda  and
      Pino, Juan  and
      Bali, Kalika",
    booktitle = "Findings of the Association for Computational Linguistics: EMNLP 2023",
    month = dec,
    year = "2023",
    address = "Singapore",
    publisher = "Association for Computational Linguistics",
    url = "https://aclanthology.org/2023.findings-emnlp.790/",
    doi = "10.18653/v1/2023.findings-emnlp.790",
    pages = "11807--11816"
}

@article{guo2022survey,
  title={A survey on automated fact-checking},
  author={Guo, Zhijiang and Schlichtkrull, Michael and Vlachos, Andreas},
  journal={Transactions of the Association for Computational Linguistics},
  volume={10},
  pages={178--206},
  year={2022},
  publisher={MIT Press One Rogers Street, Cambridge, MA 02142-1209, USA journals-info~…},
  url={https://doi.org/10.1162/tacl_a_00454}
}

@inproceedings{vo2018rise,
author = {Vo, Nguyen and Lee, Kyumin},
title = {The Rise of Guardians: Fact-checking URL Recommendation to Combat Fake News},
year = {2018},
isbn = {9781450356572},
url = {https://doi.org/10.1145/3209978.3210037},
doi = {10.1145/3209978.3210037},
booktitle = {Proceedings of the 41st International ACM SIGIR Conference on Research \& Development in Information Retrieval},
pages = {275--284}
}

@article{Kadhim2019SurveyOS,
  title={Survey on supervised machine learning techniques for automatic text classification},
  author={Ammar Ismael Kadhim},
  journal={Artificial Intelligence Review},
  year={2019},
  volume={52},
  pages={273-292},
  url={https://api.semanticscholar.org/CorpusID:58020300}
}

@article{Minaee2020DeepLT,
author = {Minaee, Shervin and Kalchbrenner, Nal and Cambria, Erik and Nikzad, Narjes and Chenaghlu, Meysam and Gao, Jianfeng},
title = {Deep Learning--based Text Classification: A Comprehensive Review},
year = {2021},
issue_date = {April 2022},
publisher = {Association for Computing Machinery},
address = {New York, NY, USA},
volume = {54},
number = {3},
issn = {0360-0300},
url = {https://doi.org/10.1145/3439726},
doi = {10.1145/3439726},
journal = {ACM Computing Survery},
month = apr,
articleno = {62},
numpages = {40}
}

@inproceedings{Wang2017LiarLP,
    title = "{\textquotedblleft}Liar, Liar Pants on Fire{\textquotedblright}: A New Benchmark Dataset for Fake News Detection",
    author = "Wang, William Yang",
    editor = "Barzilay, Regina  and
      Kan, Min-Yen",
    booktitle = "Proceedings of the 55th Annual Meeting of the Association for Computational Linguistics (Volume 2: Short Papers)",
    month = jul,
    year = "2017",
    address = "Vancouver, Canada",
    publisher = "Association for Computational Linguistics",
    url = "https://aclanthology.org/P17-2067/",
    doi = "10.18653/v1/P17-2067",
    pages = "422--426"
}

@inproceedings{Kim2014ConvolutionalNN,
    title = "Convolutional Neural Networks for Sentence Classification",
    author = "Kim, Yoon",
    editor = "Moschitti, Alessandro  and
      Pang, Bo  and
      Daelemans, Walter",
    booktitle = "Proceedings of the 2014 Conference on Empirical Methods in Natural Language Processing",
    month = oct,
    year = "2014",
    address = "Doha, Qatar",
    publisher = "Association for Computational Linguistics",
    url = "https://aclanthology.org/D14-1181/",
    doi = "10.3115/v1/D14-1181",
    pages = "1746--1751"
}

@inproceedings{Ma2016DetectingRF,
author = {Ma, Jing and Gao, Wei and Mitra, Prasenjit and Kwon, Sejeong and Jansen, Bernard J. and Wong, Kam-Fai and Cha, Meeyoung},
title = {Detecting rumors from microblogs with recurrent neural networks},
year = {2016},
isbn = {9781577357704},
booktitle = {Proceedings of the 25th International Joint Conference on Artificial Intelligence},
pages = {3818--3824},
url={https://dl.acm.org/doi/10.5555/3061053.3061153}
}

@inproceedings{ma-etal-2020-mode,
    title = "{MODE}-{LSTM}: A Parameter-efficient Recurrent Network with Multi-Scale for Sentence Classification",
    author = "Ma, Qianli  and
      Lin, Zhenxi  and
      Yan, Jiangyue  and
      Chen, Zipeng  and
      Yu, Liuhong",
    editor = "Webber, Bonnie  and
      Cohn, Trevor  and
      He, Yulan  and
      Liu, Yang",
    booktitle = "Proceedings of the 2020 Conference on Empirical Methods in Natural Language Processing",
    month = nov,
    year = "2020",
    address = "Online",
    publisher = "Association for Computational Linguistics",
    url = "https://aclanthology.org/2020.emnlp-main.544/",
    doi = "10.18653/v1/2020.emnlp-main.544",
    pages = "6705--6715"
}

@inproceedings{Sachan2019RevisitingLN,
  title={Revisiting LSTM networks for semi-supervised text classification via mixed objective function},
  author={Sachan, Devendra Singh and Zaheer, Manzil and Salakhutdinov, Ruslan},
  booktitle={Proceedings of the AAAI Conference on Artificial Intelligence},
  pages={6940--6948},
  year={2019},
  url={https://doi.org/10.1609/aaai.v33i01.33016940}
}

@inproceedings{sun-lu-2020-understanding,
    title = "Understanding Attention for Text Classification",
    author = "Sun, Xiaobing  and
      Lu, Wei",
    editor = "Jurafsky, Dan  and
      Chai, Joyce  and
      Schluter, Natalie  and
      Tetreault, Joel",
    booktitle = "Proceedings of the 58th Annual Meeting of the Association for Computational Linguistics",
    month = jul,
    year = "2020",
    address = "Online",
    publisher = "Association for Computational Linguistics",
    url = "https://aclanthology.org/2020.acl-main.312/",
    doi = "10.18653/v1/2020.acl-main.312",
    pages = "3418--3428"
}

@inproceedings{linmei-etal-2019-heterogeneous,
    title = "Heterogeneous Graph Attention Networks for Semi-supervised Short Text Classification",
    author = "Linmei, Hu  and
      Yang, Tianchi  and
      Shi, Chuan  and
      Ji, Houye  and
      Li, Xiaoli",
    editor = "Inui, Kentaro  and
      Jiang, Jing  and
      Ng, Vincent  and
      Wan, Xiaojun",
    booktitle = "Proceedings of the 2019 Conference on Empirical Methods in Natural Language Processing and the 9th International Joint Conference on Natural Language Processing",
    month = nov,
    year = "2019",
    address = "Hong Kong, China",
    publisher = "Association for Computational Linguistics",
    url = "https://aclanthology.org/D19-1488/",
    doi = "10.18653/v1/D19-1488",
    pages = "4821--4830"
}

@inproceedings{yun2023focus,
    title = "Focus on the Core: Efficient Attention via Pruned Token Compression for Document Classification",
    author = "Yun, Jungmin  and
      Kim, Mihyeon  and
      Kim, Youngbin",
    editor = "Bouamor, Houda  and
      Pino, Juan  and
      Bali, Kalika",
    booktitle = "Findings of the Association for Computational Linguistics: EMNLP 2023",
    month = dec,
    year = "2023",
    address = "Singapore",
    publisher = "Association for Computational Linguistics",
    url = "https://aclanthology.org/2023.findings-emnlp.909/",
    doi = "10.18653/v1/2023.findings-emnlp.909",
    pages = "13617--13628"
}

@inproceedings{van-nooten-daelemans-2025-jump,
    title = "Jump To Hyperspace: Comparing {E}uclidean and Hyperbolic Loss Functions for Hierarchical Multi-Label Text Classification",
    author = "Van Nooten, Jens  and
      Daelemans, Walter",
    editor = "Rambow, Owen  and
      Wanner, Leo  and
      Apidianaki, Marianna  and
      Al-Khalifa, Hend  and
      Eugenio, Barbara Di  and
      Schockaert, Steven",
    booktitle = "Proceedings of the 31st International Conference on Computational Linguistics",
    month = jan,
    year = "2025",
    address = "Abu Dhabi, UAE",
    publisher = "Association for Computational Linguistics",
    url = "https://aclanthology.org/2025.coling-main.287/",
    pages = "4260--4273"
}

@inproceedings{xiong-etal-2021-fusing,
    title = "Fusing Label Embedding into {BERT}: An Efficient Improvement for Text Classification",
    author = "Xiong, Yijin  and
      Feng, Yukun  and
      Wu, Hao  and
      Kamigaito, Hidetaka  and
      Okumura, Manabu",
    editor = "Zong, Chengqing  and
      Xia, Fei  and
      Li, Wenjie  and
      Navigli, Roberto",
    booktitle = "Findings of the Association for Computational Linguistics: ACL-IJCNLP 2021",
    month = aug,
    year = "2021",
    address = "Online",
    publisher = "Association for Computational Linguistics",
    url = "https://aclanthology.org/2021.findings-acl.152/",
    doi = "10.18653/v1/2021.findings-acl.152",
    pages = "1743--1750"
}

@inproceedings{croce-etal-2020-gan,
    title = "{GAN}-{BERT}: Generative Adversarial Learning for Robust Text Classification with a Bunch of Labeled Examples",
    author = "Croce, Danilo  and
      Castellucci, Giuseppe  and
      Basili, Roberto",
    editor = "Jurafsky, Dan  and
      Chai, Joyce  and
      Schluter, Natalie  and
      Tetreault, Joel",
    booktitle = "Proceedings of the 58th Annual Meeting of the Association for Computational Linguistics",
    month = jul,
    year = "2020",
    address = "Online",
    publisher = "Association for Computational Linguistics",
    url = "https://aclanthology.org/2020.acl-main.191/",
    doi = "10.18653/v1/2020.acl-main.191",
    pages = "2114--2119"
}

@inproceedings{10.1145/3637528.3671977,
author = {Wu, Jiaying and Guo, Jiafeng and Hooi, Bryan},
title = {Fake News in Sheep's Clothing: Robust Fake News Detection Against LLM-Empowered Style Attacks},
year = {2024},
isbn = {9798400704901},
url = {https://doi.org/10.1145/3637528.3671977},
doi = {10.1145/3637528.3671977},
booktitle = {Proceedings of the 30th ACM SIGKDD Conference on Knowledge Discovery and Data Mining},
pages = {3367--3378}
}

@article{10.1145/3714456,
author = {Haider Rizvi, Syed Mustafa and Imran, Ramsha and Mahmood, Arif},
title = {Text Classification Using Graph Convolutional Networks: A Comprehensive Survey},
year = {2025},
issue_date = {August 2025},
publisher = {Association for Computing Machinery},
address = {New York, NY, USA},
volume = {57},
number = {8},
issn = {0360-0300},
url = {https://doi.org/10.1145/3714456},
doi = {10.1145/3714456},
journal = {ACM Computing Survery},
month = mar,
articleno = {201},
numpages = {38}
}

@inproceedings{Bian2020RumorDO,
  title={Rumor detection on social media with bi-directional graph convolutional networks},
  author={Bian, Tian and Xiao, Xi and Xu, Tingyang and Zhao, Peilin and Huang, Wenbing and Rong, Yu and Huang, Junzhou},
  booktitle={Proceedings of the AAAI Conference on Artificial Intelligence},
  pages={549--556},
  year={2020},
  url={https://ojs.aaai.org/index.php/AAAI/article/view/5393}
}

@inproceedings{Ma2018RumorDO,
    title = "Rumor Detection on {T}witter with Tree-structured Recursive Neural Networks",
    author = "Ma, Jing  and
      Gao, Wei  and
      Wong, Kam-Fai",
    editor = "Gurevych, Iryna  and
      Miyao, Yusuke",
    booktitle = "Proceedings of the 56th Annual Meeting of the Association for Computational Linguistics (Volume 1: Long Papers)",
    month = jul,
    year = "2018",
    address = "Melbourne, Australia",
    publisher = "Association for Computational Linguistics",
    url = "https://aclanthology.org/P18-1184/",
    doi = "10.18653/v1/P18-1184",
    pages = "1980--1989"
}

@inproceedings{Deng_Wang_Zhu_Wang_Feng_2025,
    title={CrAM: Credibility-Aware Attention Modification in LLMs for Combating Misinformation in RAG},
    url={https://ojs.aaai.org/index.php/AAAI/article/view/34547},DOI={10.1609/aaai.v39i22.34547},
    booktitle={Proceedings of the AAAI Conference on Artificial Intelligence},
    author={Deng, Boyi and Wang, Wenjie and Zhu, Fengbin and Wang, Qifan and Feng, Fuli},
    year={2025},
    month={Apr.},
    pages={23760-23768}
}

@inproceedings{10.1145/3184558.3188731,
    author = {Zhang, Amy X. and Ranganathan, Aditya and Metz, Sarah Emlen and Appling, Scott and Sehat, Connie Moon and Gilmore, Norman and Adams, Nick B. and Vincent, Emmanuel and Lee, Jennifer and Robbins, Martin and Bice, Ed and Hawke, Sandro and Karger, David and Mina, An Xiao},
    title = {A Structured Response to Misinformation: Defining and Annotating Credibility Indicators in News Articles},
    year = {2018},
    isbn = {9781450356404},
    url = {https://doi.org/10.1145/3184558.3188731},
    doi = {10.1145/3184558.3188731},
    booktitle = {Companion Proceedings of the The Web Conference},
    pages = {603--612},
    numpages = {10}
}

@inproceedings{10.1145/3041021.3053379,
    author = {Popat, Kashyap},
    title = {Assessing the Credibility of Claims on the Web},
    year = {2017},
    isbn = {9781450349147},
    url = {https://doi.org/10.1145/3041021.3053379},
    doi = {10.1145/3041021.3053379},
    booktitle = {Proceedings of the 26th International Conference on World Wide Web Companion},
    pages = {735--739},
}

@inproceedings{10.1145/3274327,
  title={Fake cures: user-centric modeling of health misinformation in social media},
  author={Ghenai, Amira and Mejova, Yelena},
  booktitle={Proceedings of the ACM on Human-Computer Interaction},
  pages={1--20},
  year={2018},
  url = {https://doi.org/10.1145/3274327}
}

@inproceedings{teng2022characterizing,
  title={Characterizing user susceptibility to COVID-19 misinformation on Twitter},
  author={Teng, Xian and Lin, Yu-Ru and Chung, Wen-Ting and Li, Ang and Kovashka, Adriana},
  booktitle={Proceedings of the International AAAI Conference on Web and Social Media},
  pages={1005--1016},
  year={2022},
  url={https://doi.org/10.1609/icwsm.v16i1.19353}
}

@inproceedings{10.1145/3404835.3462990,
author = {Dou, Yingtong and Shu, Kai and Xia, Congying and Yu, Philip S. and Sun, Lichao},
title = {User Preference-aware Fake News Detection},
year = {2021},
isbn = {9781450380379},
url = {https://doi.org/10.1145/3404835.3462990},
doi = {10.1145/3404835.3462990},
booktitle = {Proceedings of the 44th International ACM SIGIR Conference on Research and Development in Information Retrieval},
pages = {2051--2055}
}

@inproceedings{dun2021kan,
  title={Kan: Knowledge-aware attention network for fake news detection},
  author={Dun, Yaqian and Tu, Kefei and Chen, Chen and Hou, Chunyan and Yuan, Xiaojie},
  booktitle={Proceedings of the AAAI Conference on Artificial Intelligence},
  pages={81--89},
  year={2021},
  url={https://ojs.aaai.org/index.php/AAAI/article/view/16080}
}

@inproceedings{Wang2018EANNEA,
author = {Wang, Yaqing and Ma, Fenglong and Jin, Zhiwei and Yuan, Ye and Xun, Guangxu and Jha, Kishlay and Su, Lu and Gao, Jing},
title = {EANN: Event Adversarial Neural Networks for Multi-Modal Fake News Detection},
year = {2018},
isbn = {9781450355520},
url = {https://doi.org/10.1145/3219819.3219903},
doi = {10.1145/3219819.3219903},
booktitle = {Proceedings of the 24th ACM SIGKDD International Conference on Knowledge Discovery \& Data Mining},
pages = {849--857}
}

@inproceedings{10.1145/3589334.3648152,
author = {Shang, Lanyu and Zhang, Yang and Chen, Bozhang and Zong, Ruohan and Yue, Zhenrui and Zeng, Huimin and Wei, Na and Wang, Dong},
title = {MMAdapt: A Knowledge-guided Multi-source Multi-class Domain Adaptive Framework for Early Health Misinformation Detection},
year = {2024},
isbn = {9798400701719},
url = {https://doi.org/10.1145/3589334.3648152},
doi = {10.1145/3589334.3648152},
booktitle = {Proceedings of the ACM Web Conference},
pages = {4653--4663},
}

@inproceedings{li-etal-2025-semantic,
    title = "Semantic Reshuffling with {LLM} and Heterogeneous Graph Auto-Encoder for Enhanced Rumor Detection",
    author = "Li, Guoyi  and
      Hu, Die  and
      Liu, Zongzhen  and
      Zhang, Xiaodan  and
      Lyu, Honglei",
    editor = "Rambow, Owen  and
      Wanner, Leo  and
      Apidianaki, Marianna  and
      Al-Khalifa, Hend  and
      Eugenio, Barbara Di  and
      Schockaert, Steven",
    booktitle = "Proceedings of the 31st International Conference on Computational Linguistics",
    month = jan,
    year = "2025",
    address = "Abu Dhabi, UAE",
    publisher = "Association for Computational Linguistics",
    url = "https://aclanthology.org/2025.coling-main.572/",
    pages = "8557--8572"
}

@article{thorson2010credibility,
  title={Credibility in context: How uncivil online commentary affects news credibility},
  author={Thorson, Kjerstin and Vraga, Emily and Ekdale, Brian},
  journal={Mass Communication and Society},
  volume={13},
  number={3},
  pages={289--313},
  year={2010},
  publisher={Taylor \& Francis},
  url={https://doi.org/10.1080/15205430903225571}
}

@inproceedings{guu2020retrieval,
author = {Guu, Kelvin and Lee, Kenton and Tung, Zora and Pasupat, Panupong and Chang, Ming-Wei},
title = {REALM: retrieval-augmented language model pre-training},
year = {2020},
booktitle = {Proceedings of the 37th International Conference on Machine Learning},
articleno = {368},
pages={3929--3938},
url={https://dl.acm.org/doi/abs/10.5555/3524938.3525306}
}

@inproceedings{lewis2020rag,
author = {Lewis, Patrick and Perez, Ethan and Piktus, Aleksandra and Petroni, Fabio and Karpukhin, Vladimir and Goyal, Naman and K\"{u}ttler, Heinrich and Lewis, Mike and Yih, Wen-tau and Rockt\"{a}schel, Tim and Riedel, Sebastian and Kiela, Douwe},
title = {Retrieval-augmented generation for knowledge-intensive NLP tasks},
year = {2020},
isbn = {9781713829546},
booktitle = {Proceedings of the 34th International Conference on Neural Information Processing Systems},
url = {https://dl.acm.org/doi/abs/10.5555/3495724.3496517},
pages = {9459--9474}
}

@inproceedings{perez2018film,
  title={Film: Visual reasoning with a general conditioning layer},
  author={Perez, Ethan and Strub, Florian and De Vries, Harm and Dumoulin, Vincent and Courville, Aaron},
  booktitle={Proceedings of the AAAI Conference on Artificial Intelligence},
  year={2018},
  pages = {3942--3951},
  url={https://dl.acm.org/doi/abs/10.5555/3504035.3504518}
}

@inproceedings{wang-etal-2016-learning-represent,
    title = "Learning to Represent Review with Tensor Decomposition for Spam Detection",
    author = "Wang, Xuepeng  and
      Liu, Kang  and
      He, Shizhu  and
      Zhao, Jun",
    editor = "Su, Jian  and
      Duh, Kevin  and
      Carreras, Xavier",
    booktitle = "Proceedings of the 2016 Conference on Empirical Methods in Natural Language Processing",
    month = nov,
    year = "2016",
    address = "Austin, Texas",
    publisher = "Association for Computational Linguistics",
    url = "https://aclanthology.org/D16-1083/",
    doi = "10.18653/v1/D16-1083",
    pages = "866--875"
}

@inproceedings{lu-koehn-2025-learn,
    title = "Learn and Unlearn: Addressing Misinformation in Multilingual {LLM}s",
    author = "Lu, TaiMing  and
      Koehn, Philipp",
    editor = "Christodoulopoulos, Christos  and
      Chakraborty, Tanmoy  and
      Rose, Carolyn  and
      Peng, Violet",
    booktitle = "Proceedings of the 2025 Conference on Empirical Methods in Natural Language Processing",
    month = nov,
    year = "2025",
    address = "Suzhou, China",
    publisher = "Association for Computational Linguistics",
    url = "https://aclanthology.org/2025.emnlp-main.516/",
    doi = "10.18653/v1/2025.emnlp-main.516",
    pages = "10191--10206",
    ISBN = "979-8-89176-332-6"
}

@article{touvron2023llama,
  title={Llama: Open and efficient foundation language models},
  author={Touvron, Hugo and Lavril, Thibaut and Izacard, Gautier and Martinet, Xavier and Lachaux, Marie-Anne and Lacroix, Timoth{\'e}e and Rozi{\`e}re, Baptiste and Goyal, Naman and Hambro, Eric and Azhar, Faisal and others},
  journal={arXiv preprint arXiv:2302.13971},
  year={2023},
  url={https://arxiv.org/abs/2302.13971}
}

@article{liu2019roberta,
  title={Roberta: A robustly optimized bert pretraining approach},
  author={Liu, Yinhan and Ott, Myle and Goyal, Naman and Du, Jingfei and Joshi, Mandar and Chen, Danqi and Levy, Omer and Lewis, Mike and Zettlemoyer, Luke and Stoyanov, Veselin},
  journal={arXiv preprint arXiv:1907.11692},
  year={2019},
  url={https://arxiv.org/abs/1907.11692}
}

@article{he2020deberta,
  title={Deberta: Decoding-enhanced \\ bert with disentangled attention},
  author={He, Pengcheng and Liu, Xiaodong and Gao, Jianfeng and Chen, Weizhu},
  journal={arXiv preprint arXiv:2006.03654},
  year={2020},
  url={https://arxiv.org/abs/2006.03654}
}

\clearpage
\appendix

\section{Methodology Details}
\label{sec:appendix}
\textbf{Pipeline Overview}. Given an input $x_i$, BiMind first encodes the text via a Transformer encoder with an AGA module to produce a text representation $\boldsymbol{h}$ from the content itself. In parallel, BiMind retrieves top-$k$ semantically similar examples from an in-domain memory bank and injects them into $\boldsymbol{h}$ through FiLM to form a knowledge-augmented representation $\boldsymbol{h}_E$. The two heads then perform separate predictions and generate $\boldsymbol{z}_0$ and $\boldsymbol{z}_E$, where the fusion weight $g$ is computed using an uncertainty-aware fusion strategy and $\boldsymbol{z}_0$ and $\boldsymbol{z}_E$ are combined into the final prediction $\boldsymbol{z}_F$. Finally, we quantify how much the retrieved knowledge helps by using the VoX metric.

Here, we chose these POS categories based on empirical patterns observed in our preliminary analysis, which show that certain word types, such as adjectives and pronouns, provide strong discriminative signals. Beyond improving accuracy, our AGA design makes the model's attention behavior more stable and interpretable by grounding it in linguistic patterns. A universal POS tagging framework is adopted, and the formulation is language-agnostic, with only the tagging tool needing to be adapted for different languages.

\textbf{AGA}. In the AGA module, a learnable per-head temperature $\tau_h$ is applied before softmax to normalize the logits and produce attention weights $\alpha^{(h)}_{i,j}$:
\begin{equation}
\alpha^{(h)}_{i,j}
=
\mathrm{softmax}_j\!\left(\frac{\widetilde{\boldsymbol{A}}^{(h)}_{i,j}}{\tau_h}\right)
\label{eq:aga_softmax}
\end{equation}
Each head output is then computed as a weighted sum $\boldsymbol{o}^{(h)}_i$ of values $\boldsymbol{v}^{(h)}_j$:
\begin{equation}
\boldsymbol{o}^{(h)}_i
=
\sum_{j=1}^{L} \alpha^{(h)}_{i,j}\,\boldsymbol{v}^{(h)}_j
\label{eq:aga_head_out}
\end{equation}
and the final output of multi-head attention (MHA) is: 
\begin{equation}
\mathrm{MHA}(\boldsymbol{E}(x_i))
=
\mathrm{Concat}\!\left(\boldsymbol{o}^{(1)}_i,\ldots,\boldsymbol{o}^{(H)}_i\right)\boldsymbol{W}_o
\label{eq:aga_mha}
\end{equation}
where $H$ is the number of attention heads in MHA and $\boldsymbol{W}_o$ is the weight matrix.

\textbf{Entropy-gated Fusion}. Uncertainty-aware fusion strategy is a weighting strategy that generates the final prediction based on the model's entropy. A higher entropy indicates the prediction is less confident. Here, we formulate the gate input vector $\boldsymbol{u}$ as:
\begin{equation}
    \boldsymbol{u} = [\boldsymbol{h}; \boldsymbol{h}_E; \mathcal{H}(\hat{\boldsymbol{y}}_0); \mathcal{H}(\hat{\boldsymbol{y}}_E)]
\end{equation}
where $\mathcal{H}(\hat{\boldsymbol{y}}_0)$, $\mathcal{H}(\hat{\boldsymbol{y}}_E)$ are the entropy of Head$_0$ and Head$_E$.

\textbf{Trainable Agreement Head}. In the trainable agreement head scheme, we combine two streams of features from both heads and add a new classifier to learn how to jointly leverage them, instead of directly fusing predictions. To construct agreement features, we combine:
\begin{itemize}
    \item hidden states $\boldsymbol{h}$ and $\boldsymbol{h}_E$,
    \item elementwise interaction $(\boldsymbol{h} \odot \boldsymbol{h}_E)$,
    \item and absolute difference $(|\boldsymbol{h} - \boldsymbol{h}_E|)$.
\end{itemize}
Formally, the agreement feature vector is defined as:
\begin{equation}
\boldsymbol{\phi}_{\text{agree}} = [\,\boldsymbol{h};\;\boldsymbol{h}_E;\;\boldsymbol{h} \odot \boldsymbol{h}_E;\;|\boldsymbol{h} - \boldsymbol{h}_E|\,].
\end{equation}
Then, the agreement features are fed into the MLP layers:
\begin{equation}
\boldsymbol{z}_A = \boldsymbol{W}_2 \,\sigma(\boldsymbol{W}_1 \boldsymbol{\phi}_{\text{agree}} + \boldsymbol{b}_1) + \boldsymbol{b}_2, \quad \boldsymbol{z}_A \in \mathbb{R}^K,
\end{equation}
where $\boldsymbol{z}_A$ is agreement logits. $\boldsymbol{W}_1, \boldsymbol{W}_2$ and $\boldsymbol{b}_1, \boldsymbol{b}_2$ are learnable parameters.
Finally, the agreement head outputs predictions as:
\begin{equation}
\boldsymbol{y}_A = \mathrm{softmax}(\boldsymbol{z}_A).
\end{equation}
Here, $\boldsymbol{y}_A$ is the "agreement head" prediction, which learns to explore the consistency and discrepancy between the two reasoning heads.

\section{Experimental Model}
\textbf{Model Configuration}. Our detection pipeline employs a BiMind design, which effectively integrates textual and knowledge features. Each input token is represented by a 128-dimensional embedding vector. The maximum sequence length is 5000. In the transformer-based classifier module, our model consists of 2 stacked transformer encoder layers with a multi-head attention scheme (i.e., 16 attention heads), producing pooled text representations. This custom 2-layer transformer encoder is trained from scratch as the backbone for content-internal prediction heads. Unlike our LLM variants, this architecture does not use pretrained models, and it encodes text only and learns task-specific representations end-to-end. 

Then, we construct two heads: a content-internal head that incorporates the text representations with the AGA module, and an external knowledge head that injects the knowledge vectors via FiLM before MLP. Here, we set $k$ to 3 in the knowledge retrieval module, based on SentenceTransformer \citep{reimers-2019-sentence-bert}. In the feature fusion function, we set the entropy-gated strategy as the default, where other options include a trainable agreement head, standard logit average, and product-of-experts. For the two heads, we employ ReLU as an activation function and set dropout regularization to 0.3, where both heads are trained with CE loss and a symmetric-KL agreement regularizer. To handle the class imbalance issue, we adopt class-balanced weights on the CE loss. The final output layer with a softmax function is designed to provide the probability distribution indicating the likelihood of the content being labeled 1 (i.e., correct information) or 0 (i.e., incorrect information). Here, we use Adam optimizer with learning rate $1 \times 10^{-5}$ and batch size 64 to train our model, where we employ the mixed-precision training, gradient clipping, and early-stopping (patience = 3) to tune the hyperparameters.
% Here, we employ Adam optimizer, mixed-precision training, and early stopping to effectively reach convergence.

\textbf{Experimental Models}. In the experiment setup, we compared our BiMind model with five tested models, including CNN-based model \citep{Kim2014ConvolutionalNN}, GCN-based model \citep{Yao2018GraphCN}, HAN \citep{Yang2016HierarchicalAN}, BERT \citep{devlin2019bert}, and HeteroSGT \citep{Zhang2024HeterogeneousST}. Technically, the CNN-based model employs CNN layers to extract text features from article content and then uses the extracted features to detect incorrect information. The GCN-based model explores the weighted graph built on news articles, which uses a GCN for identifying incorrect information. HAN applies word-level and sentence-level features in news content for incorrect information detection. Here, BERT is a transformer-based model, similar to our transformer-based classifier. HeteroSGT explores the heterogeneous subgraph transformer to classify articles via the heterogeneous graph.

\textbf{Model Parameters}.
In the main body of this paper, we used the LLaMA-7B model as the backbone. We reported both total and trainable parameters for all models we tested and compared the SentenceTransformer-based baseline model against several LLMs:
\begin{itemize}
    \item {Small/Medium models (110M-184M parameters)}: BERT, RoBERTa, and DeBERTa .
    \item {Large model (7B parameters)}: LLaMA-7B.
\end{itemize}
Detailed results are provided in Appendix Tables~\ref{tab:appendix_full_results_std} and~\ref{tab:runtime}. To demonstrate model scalability, we designed three versions of BiMind, as shown in Table~\ref{tab:paras}.
\begin{table}[t]
\centering
\scriptsize
\setlength{\tabcolsep}{4pt} 
\begin{tabular}{lccc}
\toprule
\textbf{Model} & \textbf{Model Size} & \textbf{Trainable Params} & \textbf{Ratio} \\
\midrule
BiMind-Custom (Tiny) & 3.5M & 3.5M & 100\% \\
BiMind-RoBERTa (Medium) & 125M & 3.3M & 2.6\% \\
BiMind-LLaMA (Large) & 7B & 44.4M & $<$1\% \\
\bottomrule
\end{tabular}
\caption{Parameter efficiency across BiMind variants.}
\label{tab:paras}
\end{table}
In this work, we can see that BiMind has good performance, not just due to the model size. Our results show:
\begin{itemize}
    \item {Model efficiency:} BiMind outperforms full-sized models like BERT even though it has far fewer parameters that need training.
    \item {Performance consistency:} Our system shows performance improvements whether we use a small, medium, or large model as the base.
\end{itemize}
To find the best settings for both the standard BiMind and the LLM-based BiMind, we use Optuna to test a set of different hyperparameter settings and keep random seeds fixed for reproducibility, such as:
\begin{itemize}
    \item For the standard BiMind: We test different model structures, including: how many layers and heads, learning rate, and dropout.
    \item For the LLM-based BiMind: Since the LLM is frozen, we focus on tuning the parts we added, such as the AGA module.
\end{itemize}
To ensure our comparison between models is fair, we follow these rules:
\begin{itemize}
    \item Training setup: Every model is evaluated using the same data split strategy and learning objectives.
    \item No peeking: We never look at the final test results when picking the best settings.
    % \item Transparency: We will release the full list of every setting we tried, the best settings we chose, and the code used to find them on GitHub.
\end{itemize}

\begin{table}[t]
  \centering
  \scriptsize
  \setlength{\tabcolsep}{4pt}  % Optional: adjusts column padding
  \renewcommand{\arraystretch}{1.2}  % Optional: row spacing
  \begin{tabular}{lrrrr}
    \toprule
    \textbf{Dataset} & \textbf{\# Label 0} & \textbf{\# Label 1} & \textbf{\# Total} & \textbf{Avg. Length (words)} \\
    \midrule
    MM COVID   & 1,888 & 1,162  & 3,048   & 25  \\
    RoCOVery   &   605 & 1,294  & 1,899   & 500 \\
    LIAR       & 2,507 & 2,053  & 4,560   & 17  \\
    MC Fake    & 2,671 &12,621  &15,292   & 300 \\
    \bottomrule
  \end{tabular}
  \caption{Statistics of the datasets used in our experiments.}
  \label{tab:dataset-stats}
\end{table}

\begin{table*}[h]
\centering
\small
\setlength{\tabcolsep}{11pt}  % Optional: adjusts column padding
\renewcommand{\arraystretch}{1.2}  % Optional: row spacing
\begin{tabular}{lcccc}
\hline
\textbf{Dataset} & \textbf{Vocab Alignment (\%)} & \textbf{Max Sim.} & \textbf{Mean Sim.} & \textbf{Flesch (helps)} \\
\hline
LIAR      & 76.55 & 0.6603±0.0924 & 0.6077±0.0843 & 37.94 \\
MM COVID  & 75.60 & 0.7586±0.1228 & 0.6870±0.1082 & 31.68 \\
MC Fake   & 79.91 & 0.8176±0.1084 & 0.7751±0.1082 & -280.47 \\
ReCOVery  & 81.68 & 0.8056±0.1071 & 0.7567±0.1026 & -585.00 \\
\hline
\end{tabular}
\caption{Statistical comparison of knowledge attributes across datasets. Vocabulary alignment measures lexical intersection between inputs and the knowledge base. Retrieval relevance is reported as cosine similarity (mean $\pm$ std). Flesch Reading Ease for the \emph{helps} category reflects the linguistic complexity of retrieved knowledge.}
\label{tab:dataset_knowledge_stats}
\end{table*}
\section{Dataset Statistics}
Here, we present the statistics of the datasets we used, listed in Table~\ref{tab:dataset-stats}.

\textbf{Dataset-level Knowledge Impact Analysis}. We conducted a comprehensive statistical analysis of knowledge impact across these datasets, as shown in Table~\ref{tab:dataset_knowledge_stats}. Based on these datasets, we observed substantial but different levels of vocabulary alignment between test instances and the knowledge bank (ranging from 75.60\% to 81.68\%), showing that retrieved knowledge is largely in-domain. But, retrieval relevance and its effects differ significantly. LIAR presents the lowest retrieval similarity (mean 0.6077), indicating weaker semantic alignment between its short claims and retrieved knowledge, which limits the efficiency of knowledge injection. In addition, MM COVID shows moderate similarity (mean 0.6870) with higher variance, revealing that knowledge retrieval is more sensitive and selective: for short and noisy social media posts, knowledge injection yields large positive VoX gains when relevant knowledge is retrieved.

In contrast, MC Fake and ReCOVery both exhibit consistently high retrieval similarity (means 0.7751 and 0.7567, respectively), suggesting that retrieval quality is not the primary bottleneck. Instead, linguistic complexity is the dominant factor: retrieved knowledge in these datasets presents extremely low Flesch Reading Ease scores, and knowledge impact varies primarily with how such dense content is integrated rather than how relevant it is. In summary, these statistical results illustrate a spectrum of knowledge integration regimes, ranging from knowledge-limited (LIAR), to retrieval-sensitive (MM COVID), and finally to complexity-dominated settings (MC Fake, ReCOVery), motivating adaptive and uncertainty-aware mechanisms for mediating the impact of external knowledge.

\section{Data Leakage}
In our pipeline, we implemented several methods to avoid the data leakage issues, such as:
    \begin{itemize}
        \item {Data split}: We follow the official splits for all the datasets, such as a standard 80/10/10 split and 10-run averaging for evaluation.
        \item {Knowledge base construction}: The knowledge base is only constructed from training data. It never has access to the test set during the retrieval process.
        \item {Knowledge dropout (0.3):} We train the model to predict outputs even when knowledge is unreliable or absent.
        \item {Uncertainty-aware fusion:} When knowledge head has high entropy (uncertainty), the gate shifts weight toward the content head.
        % \item We acknowledge that our current setup could be strengthened with additional deduplication controls. In the revised work, we will explore near-duplicate detection and retrieval diagnostics strategies to further prevent data leakage.
    \end{itemize}

Additionally, we conducted additional experiments, including duplicate check, cheating test, and leakage-aware experiment. The results are summarized as below:
\begin{itemize}
    \item {Duplicate check}: We removed near-duplicate training samples using cosine similarity $\geq$ 0.85, filtering out 178/1643 samples (10.83\%). But, the train-test similarity is still significant (mean 0.71, 22.1\% $\geq$ 0.80), showing potential leakage risk.
    \item {Cheating test}: When test samples are included in the knowledge base, retrieval becomes perfect (top-1 similarity = 1.0, 100\% $\geq$ 0.95). A KNN method explores the leakage and improves from 78.82\% to 89.66\% (+10.84\%), while our full model is unchanged (90.15\% $\rightarrow$ 90.15\%), showing it does not use leakage.
    \item {Leakage-aware experiment}: Full model performance is stable across different setups:
    \begin{itemize}
        \item Setup A (deduplicated knowledge base): 90.15\%
        \item Setup B (external Wikipedia): 90.64\%
        \item Setup C (leakage baseline): 90.15\%
    \end{itemize}
    In contrast, KNN model shows A $\approx$ 78.82\% < C $\approx$ 89.66\%, confirming that leakage is detectable but not utilized by our model.
\end{itemize}

\textbf{Empirical Results}. From the results in Table~\ref{tab:emp}, we can see that the fusion mechanism significantly mitigates negative transfer, where fused performance exceeds both individual heads. From our analysis in Table~\ref{tab:knowim}, we observed: If data leakage existed, we would expect $>$50\% "Helps" cases (i.e., the model simply retrieving answers). The 15\% improvement aligns with semantic knowledge transfer, not memorization. In addition, we measured retrieval similarity, and if the model is cheating or memorizing answers, we would see very high similarity scores (i.e., near 1.0). Instead, our results show:
    \begin{itemize}
        \item {Low similarity}: The average similarity between what the model finds and what it's searching is quite low (around 0.52), showing that it is finding related ideas, not exact copies.
        \item {Mixed results}: Though the retrieved information helps the model 15\% of the time, it actually confuses the model 8\% of the time. If the model were "leaking" the correct answers, it would be right almost 100\% of the time.
    \end{itemize}

\begin{table}[h]
\centering
\scriptsize
\setlength{\tabcolsep}{4pt} 
\begin{tabular}{lccc}
\toprule
\textbf{Metric} & \textbf{Content Head} & \textbf{Knowledge Head} & \textbf{Fused} \\
\midrule
Accuracy & 68.2\% & 61.5\% (-6.7\%) & 70.1\% (+1.9\%) \\
F1-Score & 0.65 & 0.58 & \textbf{0.68} \\
\bottomrule
\end{tabular}
\caption{Empirical results on fusion strategy.}
\label{tab:emp}
\end{table}

\begin{table}[h]
\centering
\scriptsize
\setlength{\tabcolsep}{4pt} 
\begin{tabular}{lcc}
\toprule
\textbf{Knowledge Impact} & \textbf{\% of Test Set} & \textbf{Interpretation} \\
\midrule
Helps (wrong $\rightarrow$ correct) & $\sim$15\% & Genuine knowledge benefit \\
Hurts (correct $\rightarrow$ wrong) & $\sim$8\% & Knowledge introduces noise \\
Neutral / Both Wrong & $\sim$77\% & No leaked advantage \\
\bottomrule
\end{tabular}
\caption{Knowledge impact on data leakage.}
\label{tab:knowim}
\end{table}

For future work, we plan to design and evaluate other knowledge filtering mechanisms, such as confidence-based knowledge injection or learning a classifier to predict whether retrieval knowledge will help.

% To be more careful, we will add a section to discuss how to mitigate leakage to our paper:
% \begin{itemize}
%     \item \textbf{Duplicate check:} where we will remove any training samples that are too similar to the test samples based on retrieval statistics and similarity filtering.
%     \item \textbf{Cheating test:} We will run a test where we let the model see the test set to show what ``cheating'' looks like, so readers can see that our results are much more accurate without seeing any test data.
% \end{itemize}

% To completely address this concern, we will conduct a leakage-aware retrieval experiment:
% \begin{itemize}
%     \item Setup A: In-domain KB with deduplication
%     \item Setup B: Wikipedia data (external corpus)
%     \item Setup C: Include test set in KB (intentional leakage baseline)

%     Expected outcome: Setup A $\approx$ Setup B $<$ Setup C, validating that current gains are not driven by leakage.
% \end{itemize}

\begin{table*}[t]
\centering
\small
\setlength{\tabcolsep}{7pt}
\renewcommand{\arraystretch}{1.5}
\begin{tabular}{lllcccc}
\hline
\textbf{Model} & \textbf{Dataset} & \textbf{Head/Mode} & \textbf{Acc} & \textbf{F1} & \textbf{Pre} & \textbf{Rec} \\
\hline
\multirow{9}{*}{LLaMA-7B}
& \multirow{3}{*}{ReCOVery}
& $\text{Head}_0$ & 91.13 $\pm$ 2.34 & 91.03 $\pm$ 2.59 & 91.35 $\pm$ 2.32 & 91.13 $\pm$ 2.34 \\
& & $\text{Head}_E$ & 91.04 $\pm$ 1.18 & 91.09 $\pm$ 1.21 & 91.53 $\pm$ 1.44 & 91.04 $\pm$ 1.18 \\
& & Fused & \textbf{91.82 $\pm$ 1.29} & \textbf{91.86 $\pm$ 1.27} & \textbf{92.20 $\pm$ 1.26} & \textbf{91.82 $\pm$ 1.29} \\

& \multirow{3}{*}{MM COVID}
& $\text{Head}_0$ & 94.69 $\pm$ 1.01 & 94.70 $\pm$ 1.01 & 94.46 $\pm$ 1.45 & 94.67 $\pm$ 0.99 \\
& & $\text{Head}_E$ & 94.80 $\pm$ 0.77 & 94.81 $\pm$ 0.76 & 94.48 $\pm$ 1.21 & 94.61 $\pm$ 0.64 \\
& & Fused & \textbf{95.12 $\pm$ 0.78} & \textbf{95.12 $\pm$ 0.78} & \textbf{94.98 $\pm$ 1.06} & \textbf{95.08 $\pm$ 0.86} \\

& \multirow{3}{*}{LIAR}
& $\text{Head}_0$ & \textbf{63.26 $\pm$ 0.13} & \textbf{63.26 $\pm$ 0.19} & 63.70 $\pm$ 0.20 & 63.60 $\pm$ 0.30 \\
& & $\text{Head}_E$ & 62.70 $\pm$ 0.41 & 62.42 $\pm$ 0.53 & \textbf{64.00 $\pm$ 0.20} & 63.60 $\pm$ 0.30 \\
& & Fused & 62.93 $\pm$ 0.29 & 62.89 $\pm$ 0.29 & 63.80 $\pm$ 0.20 & \textbf{63.70 $\pm$ 0.20} \\

\hline

\multirow{9}{*}{DeBERTa-v3}
& \multirow{3}{*}{ReCOVery}
& $\text{Head}_0$ & 81.38 $\pm$ 2.95 & 81.64 $\pm$ 2.71 & 83.85 $\pm$ 1.80 & 81.38 $\pm$ 2.70 \\
& & $\text{Head}_E$ & 85.72 $\pm$ 1.43 & 85.92 $\pm$ 1.25 & 86.39 $\pm$ 1.30 & 85.32 $\pm$ 1.30 \\
& & Fused & \textbf{85.81 $\pm$ 1.27} & \textbf{85.94 $\pm$ 1.16} & \textbf{86.42 $\pm$ 1.20} & \textbf{85.81 $\pm$ 0.90} \\

& \multirow{3}{*}{MM COVID}
& $\text{Head}_0$ & 93.00 $\pm$ 1.23 & 92.98 $\pm$ 1.24 & 93.07 $\pm$ 1.20 & 92.73 $\pm$ 1.40 \\
& & $\text{Head}_E$ & \textbf{94.59 $\pm$ 1.05} & \textbf{94.59 $\pm$ 1.05} & \textbf{94.47 $\pm$ 1.20} & \textbf{94.56 $\pm$ 1.00} \\
& & Fused & 94.53 $\pm$ 1.09 & 94.54 $\pm$ 1.08 & 94.43 $\pm$ 1.30 & 94.47 $\pm$ 1.10 \\

& \multirow{3}{*}{LIAR}
& $\text{Head}_0$ & 59.75 $\pm$ 0.89 & 59.14 $\pm$ 1.21 & 61.70 $\pm$ 0.50 & 60.80 $\pm$ 0.50 \\
& & $\text{Head}_E$ & \textbf{62.05 $\pm$ 0.77} & \textbf{62.02 $\pm$ 0.94} & \textbf{62.50 $\pm$ 0.50} & \textbf{62.50 $\pm$ 0.60} \\
& & Fused & 61.91 $\pm$ 0.78 & 61.87 $\pm$ 1.00 & 62.50 $\pm$ 0.50 & 62.40 $\pm$ 0.60 \\

\hline

\multirow{9}{*}{RoBERTa}
& \multirow{3}{*}{ReCOVery}
& $\text{Head}_0$ & 81.28 $\pm$ 2.96 & 81.67 $\pm$ 2.60 & 80.36 $\pm$ 2.20 & 82.84 $\pm$ 1.74 \\
& & $\text{Head}_E$ & 84.53 $\pm$ 2.10 & 84.71 $\pm$ 2.19 & 83.97 $\pm$ 1.16 & 85.76 $\pm$ 0.74 \\
& & Fused & \textbf{84.93 $\pm$ 2.45} & \textbf{85.12 $\pm$ 2.54} & \textbf{83.92 $\pm$ 1.76} & \textbf{86.13 $\pm$ 1.06} \\

& \multirow{3}{*}{MM COVID}
& $\text{Head}_0$ & 91.87 $\pm$ 1.23 & 91.88 $\pm$ 1.23 & 92.25 $\pm$ 1.44 & 91.87 $\pm$ 1.23 \\
& & $\text{Head}_E$ & \textbf{94.49 $\pm$ 0.26} & \textbf{94.49 $\pm$ 0.27} & \textbf{94.59 $\pm$ 0.26} & \textbf{94.49 $\pm$ 0.26} \\
& & Fused & 94.21 $\pm$ 0.44 & 94.22 $\pm$ 0.44 & 94.40 $\pm$ 0.55 & 94.21 $\pm$ 0.44 \\

& \multirow{3}{*}{LIAR}
& $\text{Head}_0$ & 61.18 $\pm$ 0.77 & 60.83 $\pm$ 0.83 & 62.37 $\pm$ 0.17 & 61.91 $\pm$ 0.25 \\
& & $\text{Head}_E$ & \textbf{61.88 $\pm$ 0.63} & \textbf{61.84 $\pm$ 0.73} & \textbf{62.71 $\pm$ 0.25} & \textbf{62.55 $\pm$ 0.34} \\
& & Fused & 61.83 $\pm$ 0.66 & 61.68 $\pm$ 0.78 & 62.67 $\pm$ 0.29 & 62.47 $\pm$ 0.41 \\

\hline
\end{tabular}
\caption{Extended performance comparison (mean $\pm$ std, in \%, across 10 runs, best in \textbf{bold}).}
\label{tab:appendix_full_results_std}
\end{table*}
\begin{table*}[!h]
\centering
\small
\setlength{\tabcolsep}{4pt}
\renewcommand{\arraystretch}{1.25}
\begin{tabular}{l|l|cccc|ccc}
\toprule
\textbf{Dataset} & \textbf{Model} 
& Acc & Pre & Rec & F1 
& Training & Testing & Graph\\
\midrule
\multirow{2}{*}{MM COVID}
& HeteroSGT 
& 0.915±0.009 & 0.905±0.011 & 0.883±0.013 & 0.893±0.011
& 55.11 & -- & 13.62\\
& BiMind 
& \textbf{0.902±0.116} & \textbf{0.902±0.110} & \textbf{0.898±0.142} & \textbf{0.900±0.132}
& 15.19 & 0.08 & --\\
\midrule
\multirow{2}{*}{ReCOVery}
& HeteroSGT 
& 0.727±0.023 & 0.731±0.047 & 0.585±0.036 & 0.571±0.049
& 21.94 & -- & 9.03\\
& BiMind 
& \textbf{0.879±0.017} & \textbf{0.862±0.028} & \textbf{0.843±0.203} & \textbf{0.854±0.208}
& 14.99 & 0.10 & --\\
\midrule
\multirow{2}{*}{MC Fake}
& HeteroSGT 
& 0.883±0.002 & 0.812±0.003 & 0.762±0.002 & 0.783±0.003
& 478.53 & -- & 40.04\\
& BiMind 
& \textbf{0.887±0.051} & \textbf{0.827±0.016} & \textbf{0.700±0.099} & \textbf{0.798±0.109}
& 153.01 & 0.45 & --\\
\midrule
\multirow{2}{*}{LIAR}
& HeteroSGT 
& 0.581±0.002 & 0.580±0.003 & 0.575±0.002 & 0.571±0.003
& 116.22 & -- & 14.48\\
& BiMind 
& \textbf{0.605±0.041} & \textbf{0.601±0.045} & \textbf{0.595±0.037} & \textbf{0.595±0.037}
& 73.51 & 0.52 & --\\
\bottomrule
\end{tabular}
\caption{Performance and runtime comparison between \textbf{BiMind} and \textbf{HeteroSGT} across four benchmark datasets. 
Performance is reported as mean $\pm$ std. Runtime is measured in seconds per run.}
\label{tab:runtime}
\end{table*}
\section{Experimental Results}
For the five comparison models, CNN has poor performance on all the datasets, which may result from its fixed convolutional kernels. Due to these kernels focusing on local features, the global features or dependencies might not be effectively explored in news articles and social contexts. GCN presents different results across multiple datasets and receives better detection accuracy on the MC Fake dataset. In addition, HAN and BERT are transformer-based models with attention mechanisms, and thus, the performance is comparable between them. Though HeteroSGT achieves optimal results, such as Rec and F1 on MC Fake due to its subgraph structure, it still drops performance by 19.1$\%$ on Acc and 19.1$\%$ on Pre, 5.2$\%$ on Acc and 5.7$\%$ on Pre, respectively, compared to our proposed model on ReCOVery and LIAR datasets. Typically, on the LIAR dataset, our model achieves consistent performance across seeds, with a low standard deviation ($\pm$0.001).

\textbf{Extended Experiments}. Here, we extended the dual-head design to other LLMs, i.e., RoBERTa \citep{liu2019roberta} and DeBERTa \citep{he2020deberta}. From Table~\ref{tab:appendix_full_results_std}, we can see that separating content-internal reasoning ($\text{Head}_0$) from the knowledge-augmented reasoning ($\text{Head}_E$) shows significant dataset-relevant behavior. On knowledge-aligned datasets, like ReCOVery and MM COVID, $\text{Head}_E$ consistently improves recall and F1 score, suggesting that external knowledge provides complementary contextual signals beyond textual content alone. In contrast, on the LIAR dataset with short claims and weak retrieval alignment, $\text{Head}_E$ is not generally helpful, supporting our motivation to disentangle content inference from knowledge-based reasoning rather than enforcing unconditional knowledge injection. Additionally, the uncertainty-aware fusion strategy achieves either the best or second-best performance across models and datasets. Notably, it reduces variance and receives gains when one head performs poorly, especially on the LIAR dataset. These results validate our design choice to treat knowledge as an auxiliary, selectively trusted signal, with a fusion strategy adapting to instance-level uncertainty rather than relying on static feature concatenation alone. 

Together, these results demonstrate that the effective and reliable knowledge injection (i) conditions on data-inherent attributes, including vocabulary alignment, retrieval relevance, and sample-level linguistic complexity, and (ii) requires a principled fusion mechanism with uncertainty and agreement measurement.
%The observed improvements under the fused configuration confirm that our dual-head framework enables dynamic arbitration between textual evidence and retrieved knowledge, avoiding performance degradation caused by noisy or weakly relevant external information.

\section{Ablation Study}
When removing the AGA, it leads to a significant drop in accuracy (0.897 $\rightarrow$ 0.872) and F1 (0.895 $\rightarrow$ 0.870), showing the importance of reshaping attention logits to prevent attention collapse. Without the knowledge retrieval function, it also reduces the performance, such as a larger drop in accuracy and recall (0.897 $\rightarrow$ 0.847), indicating the significance of semantic knowledge neighbors in grounding short or ambiguous content. Additionally, replacing the uncertainty-aware fusions with a simple logit average, it causes performance degradation in F1 (0.895 $\rightarrow$ 0.861 or 0.867), showing that our fusion strategies help the model adaptively trust knowledge-augmented predictions when content-internal predictions are uncertain. Finally, removing the symmetric KL regularizer, it reduces F1 from 0.895 to 0.874, demonstrating that agreement between heads stabilizes training and improves predictions.

In conclusion, ablation results show that each component contributes complementary benefits: content features build strong baselines, attention geometries sharpen token-level salience, knowledge retrieval contextualizes content, and uncertainty fusion with an agreement regularizer ensures robust integration. In a modular and structured way, these modules jointly enable BiMind to achieve both competitive performance and interpretable diagnostics on when and why knowledge matters.

\section{Retrieval-based Methods}
We tested our model against three retrieval-based methods on ReCOVery and provided detailed comparison results in Table~\ref{tab:retr}. All models are tested using the same data and experiment settings. Below are the key findings:
\begin{itemize}
    \item {Retrieval is insufficient}: The simple retrieval-based methods (like kNN) cannot reach expected performance, showing how retrieved knowledge is used matters more than retrieval itself.
    \item {Simple fusion doesn't work:} The RAG model performs slightly worse than kNN, suggesting that just stacking knowledge does not help the model learn better.
    \item {BiMind is significantly better:} Our model is nearly 4\% more accurate than the baselines, indicating that BiMind correctly predicts a larger number of cases where other models failed.
\end{itemize}
\begin{table}[t]
\centering
\small
\setlength{\tabcolsep}{4pt}
\begin{tabular}{lccc}
\toprule
\textbf{Method} & \textbf{Accuracy} & \textbf{F1} & \textbf{Improvement} \\
\midrule
kNN (k=5) & 83.74\% & 83.54\% & - \\
kNN-BERT (k=5) & 83.74\% & 83.54\% & - \\
Standard RAG (k=3) & 83.25\% & 83.00\% & -0.49\% \\
\textbf{BiMind (Ours)} & \textbf{87.70\%} & \textbf{87.90\%} & \textbf{+3.96\%} \\
\bottomrule
\end{tabular}
\caption{Comparison with retrieval-based baselines on ReCOVery.}
\label{tab:retr}
\end{table}
\textbf{Why BiMind Wins.} Our dual-head design is more effective because:
\begin{itemize}
    \item {Two separate paths:} BiMind has one branch for the text itself and one for external knowledge, allowing it to learn which head to trust based on input features.
    \item {Conditional fusion using gating:} Instead of treating all external knowledge as equally important, our model uses a gating strategy to dynamically weight retrieved knowledge, especially when knowledge is noisy.
    \item {Knowledge dropout during training:} We train the model to handle cases where knowledge is unreliable, improving robustness.
\end{itemize}
\textbf{Knowledge Safeguards}. Here, BiMind explores multiple safeguards across retrieval, training, and architecture, such as knowledge dropout (e.g., 0.3) and an uncertainty-aware fusion strategy to prevent noisy or irrelevant retrieval from hurting performance in low-resource or short-text scenarios. Additional safeguards are summarized as follows:
\begin{itemize}
    \item Similarity-aware knowledge vector: We augment the knowledge vector with similarity signals, enabling the model to down-weight low-quality retrieval.
    \item Bounded FiLM injection: FiLM injection is bounded via $tanh$ to prevent feature distortion from noisy inputs.
    \item Agreement loss: An agreement loss regularizes divergence between reasoning heads, stabilizing learning under noisy retrieval.
    \item Leakage-aware evaluation: The leakage evaluation is performed with similarity-based deduplication (threshold = 0.85) and a cheating test to ensure robustness is not driven by retrieval leakage.
\end{itemize}
In addition, we evaluated the safeguard's robustness using the Knowledge Rejection Rate (KRR) and the Fusion Recovery Rate (FRR). When the gating mechanism does not reject irrelevant knowledge (KRR = 0.0\%), the model recovers correct predictions in 68.6\% of irrelevant cases, demonstrating strong robustness. Further analysis shows that recovery is maximized at moderate gate values (FRR = 81.8\%, gate values $\approx$ 0.44-0.45), indicating that performance relies on balancing internal and external knowledge rather than hard filtering. Additionally, retrieval similarity is not a reliable indicator of correctness (FRR: 76.9\% (medium similarity) vs. 66.7\% (high similarity)), as high-similarity neighbors can still hurt the predictions. These results highlight the importance of uncertainty-aware fusion over retrieval-only strategies.

% From experimental results on LIAR, we can see that while the knowledge head alone degrades performance, the fused model can calibrate the predictions and outperform it, indicating that the uncertainty-aware fusion strategy effectively mitigates negative knowledge impact to a meaningful extent.

To further validate our findings with RAG, we also experimented with LLaMA-7B as the backbone and compared our BiMind to LLaMA-based RAG. Even though LLaMA is a big model and improves over standard RAG (85-86\% range), BiMind still surpasses it by 1-2\%. These results show that BiMind does not just retrieval knowlegde; it can incorporate all the information from content itself and external knowledge in a structured manner. An important finding from our dual-head design is that knowledge contribution varies across instances. By analyzing the learned fusion gates, we observed that:
\begin{itemize}
    \item For easy cases (high content clarity), content head dominates (gate $\approx$ 0.8-0.9).
    \item For ambiguous cases requiring context, the knowledge head contributes more (gate $\approx$ 0.4-0.6).
\end{itemize}
From these findings, we can see that this adaptive behavior is absent in fixed-weight RAG approaches. This additional ablation study shows that removing the gating mechanism (i.e., fixed 50-50 fusion) reduces accuracy by 2.5\%, confirming that dynamic fusion is essential.

\begin{figure}[t]
    \centering
    \includegraphics[width=1.0\linewidth]{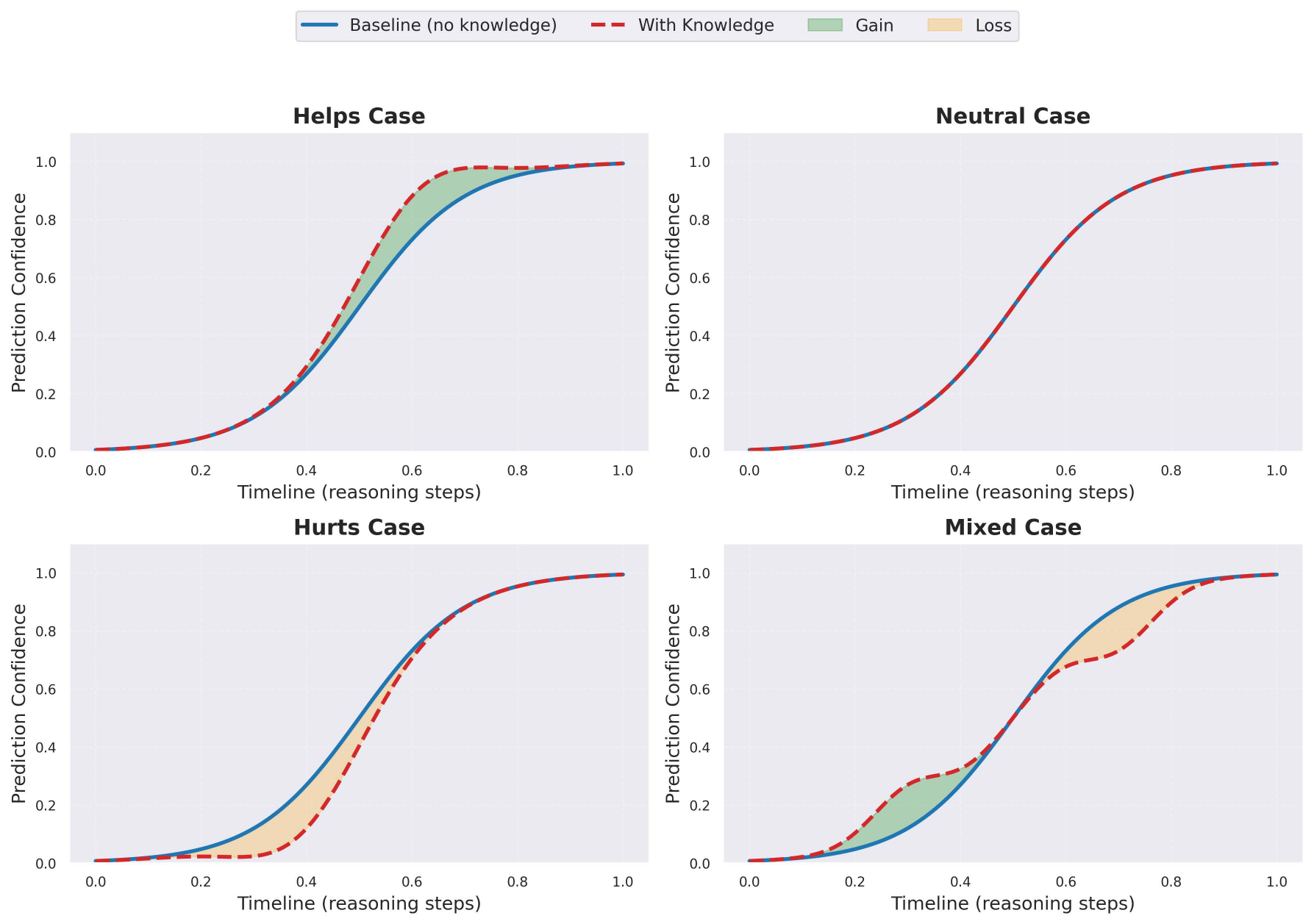}
    \caption{An illustration of knowledge impacts at the instance level. Blue curve $=$ baseline (no knowledge), red dashed curve $=$ with knowledge, shaded regions indicate gain (green) or loss (orange).}
    \label{fig:bimind_vox_cases}
\end{figure}
\section{VoX Analysis}
To further interpret the VoX values, we visualized four types of knowledge impacts in Figure~\ref{fig:bimind_vox_cases}, which demonstrates how knowledge can impact prediction confidence over the reasoning path. Typically, knowledge can help (e.g., MM COVID), be neutral, hurt (e.g., LIAR), or produce mixed patterns (e.g., MC Fake) from the dataset-level outcomes. To validate VoX interpretability, we conducted additional experiments using two metrics: $\Delta_{margin}$ and $\Delta_{entropy}$. From our results, we observed that VoX is strongly correlated with $\Delta_{margin}$ (Pearson r = 0.54, p < 0.001) and weakly correlated with $\Delta_{entropy}$ (r = 0.02), showing that higher VoX values are related to larger margin improvements and reduced uncertainty. It demonstrates that VoX represents model confidence shifts rather than calibration impact. In addition, $\Delta_{confidence}$ shows stronger alignment with performance gain ($\rho$ = 0.353, p < 0.001) and demonstrates very high agreement with VoX itself ($\rho$ = 0.754, p < 1e-38), showing that VoX effectively captures confidence differentials rather than random calibration noise. Spearman correlation between VoX and per-instance probability improvement is $\rho$ = 0.23 for gain and $\rho$ = 0.41 for correctness (both p < 0.001), statistically supporting its diagnostic validity.

\section{Quantitative Analysis}
Here, we describe the running time comparison of our BiMind framework with backbone SentenceTransformer and HeteroSGT.
%in the application of incorrect information detection.
% \begin{table}[h]
% \centering
% \scriptsize
% \setlength{\tabcolsep}{4pt}
% \renewcommand{\arraystretch}{1.2}
% \caption{Runtime comparison between BiMind and HeteroSGT across four benchmark datasets. All results are reported in seconds per run.}
% \begin{tabular}{l|ccc|cc}
% \toprule
% \multirow{2}{*}{\textbf{Dataset}} & \multicolumn{3}{c|}{\textbf{BiMind}} & \multicolumn{2}{c}{\textbf{HeteroSGT}} \\
% & Graph & Training & Testing & Graph & Training + Testing \\
% \midrule
% ReCOVery   & N/A & 14.99 & 0.10 & 9.03 & 21.94 \\
% MM COVID  & N/A & 15.19 & 0.08 & 13.62 & 55.11 \\
% LIAR       & N/A & 73.51 & 0.52 & 14.58 & 116.22 \\
% MC Fake & N/A & 153.01 & 0.45 & 40.04 & 478.53 \\
% \bottomrule
% \end{tabular}
% \label{tab:runtime}
% \end{table}
Beyond superior detection accuracy, we compared the runtime of BiMind against HeteroSGT, as shown in Table~\ref{tab:runtime}. In our framework, we skip the graph construction phase, resulting in training and testing that is nearly 4$\times$ faster (e.g, on MM COVID). HeteroSGT requires additional graph construction time (e.g., 40.04s on MC Fake) and retraining to adapt to new topics; but, BiMind generalizes with lightweight attention signals and knowledge features, showing BiMind's efficiency and scalability merits in incorrect information detection applications.

% All 32 heads exhibit nearly identical POS attention patterns, indicating late-layer representational collapse and loss of head-level specialization. Attention is selectively concentrated into a small number of heads aligned with behavior-relevant POS categories, yielding structured consolidation rather than uniform collapse.
\begin{figure}[t]
    \centering
    \includegraphics[width=\linewidth]{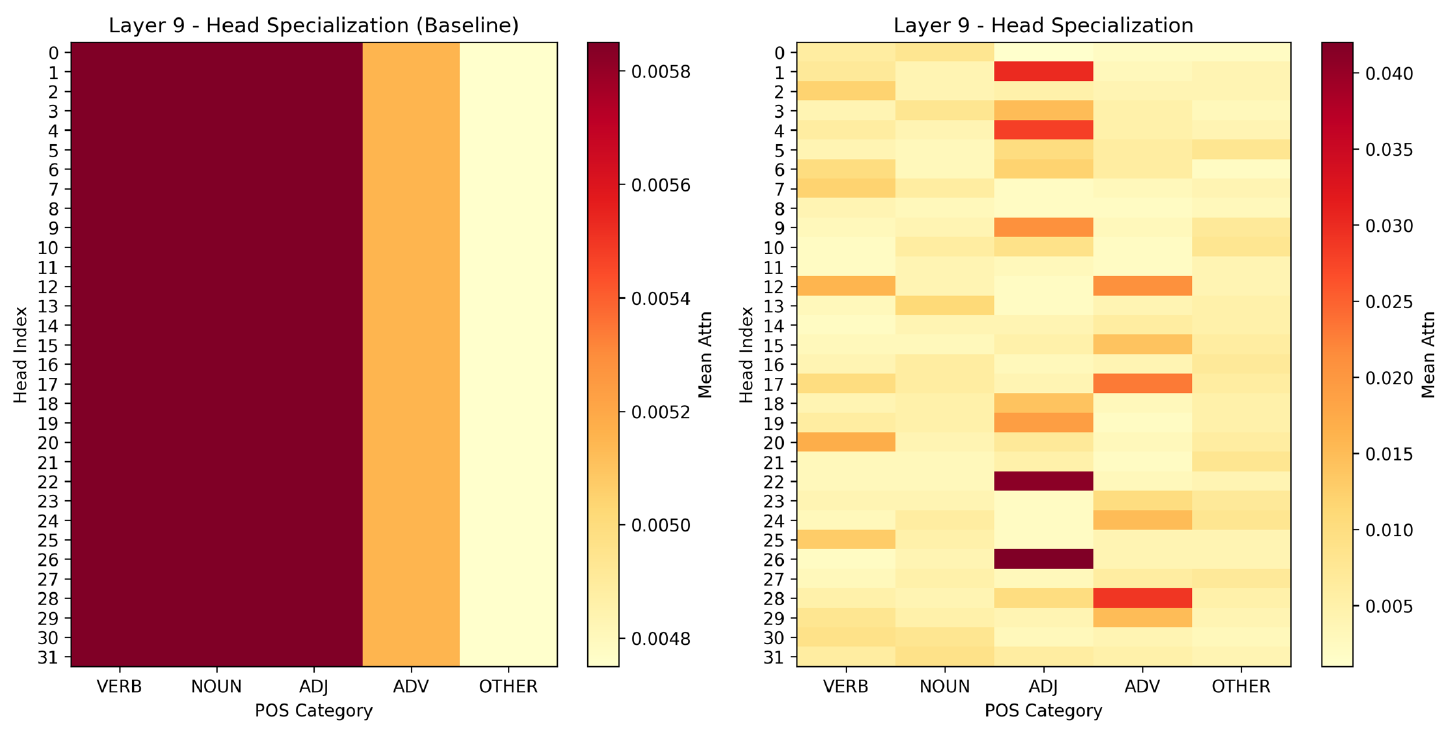}
    \caption{Layer 9 attention head specialization with and without AGA. Left: Baseline Transformer without AGA. Right: Transformer with AGA.}
    \label{fig:head-spec-layer9}
\end{figure}
\section{Attention Head Specialization}
Figure~\ref{fig:head-spec-layer9} compares attention head specialization at Layer 9 of the Transformer with and without AGA. For the baseline model, the attention heads present severe representational collapse: all heads have nearly identical attention patterns, with uniformly high focus on several categories (VERB, NOUN, ADJ) and minimal head-level variance. It shows that, without AGA, the self-attention mechanism tends to flatten linguistic structure in deeper layers.
% As a result, head identity becomes functionally irrelevant, and linguistic behavior distinctions are largely homogenized.
% geometric behavior
However, with AGA, it shows a significantly different geometric pattern. The number of active heads is reduced, but the head specialization is selectively preserved where a small number of heads focus on different categories (like ADJ and ADV). Specifically, head specialization remains diverse rather than uniform. It demonstrates that AGA transforms attention collapse into geometry-aware concentration, compressing distributed signals into a low-rank but structured representation.

\end{document}